\documentclass[review]{elsarticle}

\usepackage{amssymb}
\usepackage{amsmath}
\usepackage{booktabs}
\usepackage{multirow}
\usepackage{float}
\usepackage{comment}

\usepackage[numbers]{natbib}   

\usepackage[colorlinks=true, citecolor=blue, linkcolor=black, urlcolor=blue]{hyperref}

\journal{Information Fusion}

\begin{document}

\begin{frontmatter}

\title{Cross-Modal Mapping and Dual-Branch Reconstruction for 2D–3D Multimodal Industrial Anomaly Detection}

\author[isasi]{Radia Daci}
\ead{radiadaci@cnr.it}

\author[stiima]{Vito Ren\'o, Cosimo Patruno, Angelo Cardellicchio}
\ead{\{vito.reno|cosimo.patruno|angelo.cardellicchio\}@cnr.it}

\author[IEMN]{Abdelmalik Taleb-Ahmed}
\ead{Abdelmalik.Taleb-Ahmed@uphf.fr}

\author[isasi]{Marco Leo}
\ead{marco.leo@cnr.it}
\author[isasi]{Cosimo Distante}
\ead{cosimo.distante@cnr.it}

\affiliation[isasi]{
    organization={CNR-ISASI, Institute of Applied Sciences and Intelligent Systems / National Research Council of Italy},
    addressline={via Monteroni snc},
    city={Lecce},
    postcode={73100},
    country={Italy} 
}

\affiliation[stiima]{
    organization={CNR-STIIMA, Institute of Intelligent Industrial Technologies and Systems for Advanced Manufacturing},
    addressline={Via G. Amendola 122 D/O}, 
    city={Bari},
    postcode={70126}, 
    country={Italy}
}

\affiliation[IEMN]{
    organization={Institute d'Electronique de Microelectronique et de Nanotechnologie (IEMN) / Universite Polytechnique Hauts de France; Universite de Lille; CNRS},
    addressline={UMR 8520},
    city={City},
    postcode={59313},
    country={France}
}

\begin{abstract}
Multimodal industrial anomaly detection benefits from integrating RGB appearance with 3D surface geometry, yet existing \emph{unsupervised} approaches commonly rely on memory banks, teacher--student architectures, or fragile fusion schemes, limiting robustness under noisy depth, weak texture, or missing modalities. This paper introduces \textbf{CMDR--IAD}, a lightweight and modality-flexible unsupervised framework for reliable anomaly detection in 2D+3D multimodal as well as single-modality (2D-only or 3D-only) settings. \textbf{CMDR--IAD} combines bidirectional 2D$\leftrightarrow$3D cross-modal mapping to model appearance--geometry consistency with dual-branch reconstruction that independently captures normal texture and geometric structure. A two-part fusion strategy integrates these cues: a reliability-gated mapping anomaly highlights spatially consistent texture--geometry discrepancies, while a confidence-weighted reconstruction anomaly adaptively balances appearance and geometric deviations, yielding stable and precise anomaly localization even in depth-sparse or low-texture regions.

\noindent On the MVTec 3D-AD benchmark, CMDR--IAD achieves state-of-the-art performance while operating without memory banks, reaching 97.3\% image-level AUROC (I-AUROC), 99.6\% pixel-level AUROC (P-AUROC), and 97.6\% AUPRO. On a real-world polyurethane cutting dataset, the 3D-only variant attains 92.6\% I-AUROC and 92.5\% P-AUROC, demonstrating strong effectiveness under practical industrial conditions. These results highlight the framework's robustness, modality flexibility, and the effectiveness of the proposed fusion strategies for industrial visual inspection.
Our source code is available at \href{https://github.com/ECGAI-Research/CMDR-IAD/}{https://github.com/ECGAI-Research/CMDR-IAD/}.

\end{abstract}


\end{frontmatter}

\section{Introduction}
\label{sec:intro}

 Industrial anomaly detection (IAD) aims to identify defective or atypical products during or after manufacturing, and is therefore a key component of modern quality control pipelines. In practice, collecting representative defective samples is difficult due to their rarity, variability, and the cost of labeling. As a result, most recent methods adopt an unsupervised or one-class setting, where models are trained only on normal (nominal) data and anomalies are detected as deviations from learned regular patterns. While deep learning has significantly advanced anomaly detection in two-dimensional (2D) images, many industrial defects remain challenging to recognize from RGB appearance alone. Variations in illumination, specular reflections, or sensor noise can degrade the reliability of purely image-based approaches, and some defects manifest predominantly as subtle geometric deviations rather than as obvious color or texture changes.
 To address these challenges, recent research has increasingly turned to multimodal anomaly detection that combines RGB images with three-dimensional (3D) information such as point clouds or depth maps. 3D measurements are largely invariant to lighting and provide explicit shape and surface information, making them highly complementary to RGB signals. The introduction of benchmarks such as MVTec 3D--AD~\cite{bergmann2021mvtec} has accelerated this line of work by offering pixel-registered RGB and 3D data, fostering a variety of multimodal anomaly detection methods. 
 
\noindent Early multimodal industrial anomaly detection methods, such as {BTF} and {M3DM}, extend successful 2D approaches by modeling normality through \emph{memory banks of multimodal features}, achieving strong performance at the cost of high memory usage and limited inference speed~\cite{horwitz2023back,wang2023multimodal}. To improve efficiency, {AST} adopts a teacher--student architecture, but treats 3D information indirectly, which limits the explicit exploitation of geometric structure~\cite{rudolph2023asymmetric}. {CFM} explicitly models cross-modal feature consistency between RGB and 3D data, highlighting the importance of appearance--geometry relationships, yet remains constrained by its multimodal-only assumption~\cite{costanzino2024multimodal}. More recent approaches address robustness through cross-modal distillation, joint teacher--student memory mechanisms, and fusion architecture analysis~\cite{sui2025incomplete,liu2025multimodal,long2025revisiting}, but often introduce architectural complexity or rely on fixed fusion strategies. Existing methods exhibit clear trade-offs between segmentation performance, inference speed, and memory consumption, motivating a lightweight and modality-flexible framework with adaptive fusion.

\noindent In this work, we propose a different perspective on multimodal industrial anomaly detection based on explicit cross-modal feature mapping and dual-branch reconstruction. We introduce \textbf{CMDR--IAD} (Cross-Modal Mapping and Dual-Branch Reconstruction for 2D--3D Multimodal Industrial Anomaly Detection), an {unsupervised and modality-flexible} framework that directly models relationships between RGB-based and 3D-based features while also learning modality-specific reconstructions of normal appearance and geometry. \textbf{CMDR--IAD} employs frozen pretrained encoders for RGB images and point clouds to extract pixel-aligned feature representations, and learns two lightweight mapping networks that project 2D features into the 3D feature space and vice versa. These mappings are trained exclusively on nominal data, capturing cross-modal correspondences characteristic of defect-free samples. In parallel, two independent reconstruction decoders are trained to reproduce normal 2D and 3D features, providing modality-specific cues about deviations from normal patterns.

\noindent At inference time, \textbf{CMDR--IAD} combines four complementary signals: 2D reconstruction error, 3D reconstruction error, and two cross-modal mapping discrepancies. An adaptive fusion mechanism converts reconstruction errors into spatial reliability weights and modulates cross-modal discrepancies through a spatially varying gating function. This design allows the model to emphasize the most informative modality at each pixel and suppress noisy responses in regions with unreliable depth or weak visual texture, producing accurate and stable anomaly maps. Importantly, the use of frozen encoders and lightweight mapping and decoding networks results in competitive memory consumption and inference speed, making \textbf{CMDR--IAD} suitable for deployment in real-world industrial inspection systems. Moreover, the 3D reconstruction branch can operate independently in scenarios where only point clouds are available, as demonstrated on a real-world polyurethane cutting dataset.

\begin{figure}[h]
     \centering
    \includegraphics[width=0.7\textwidth]{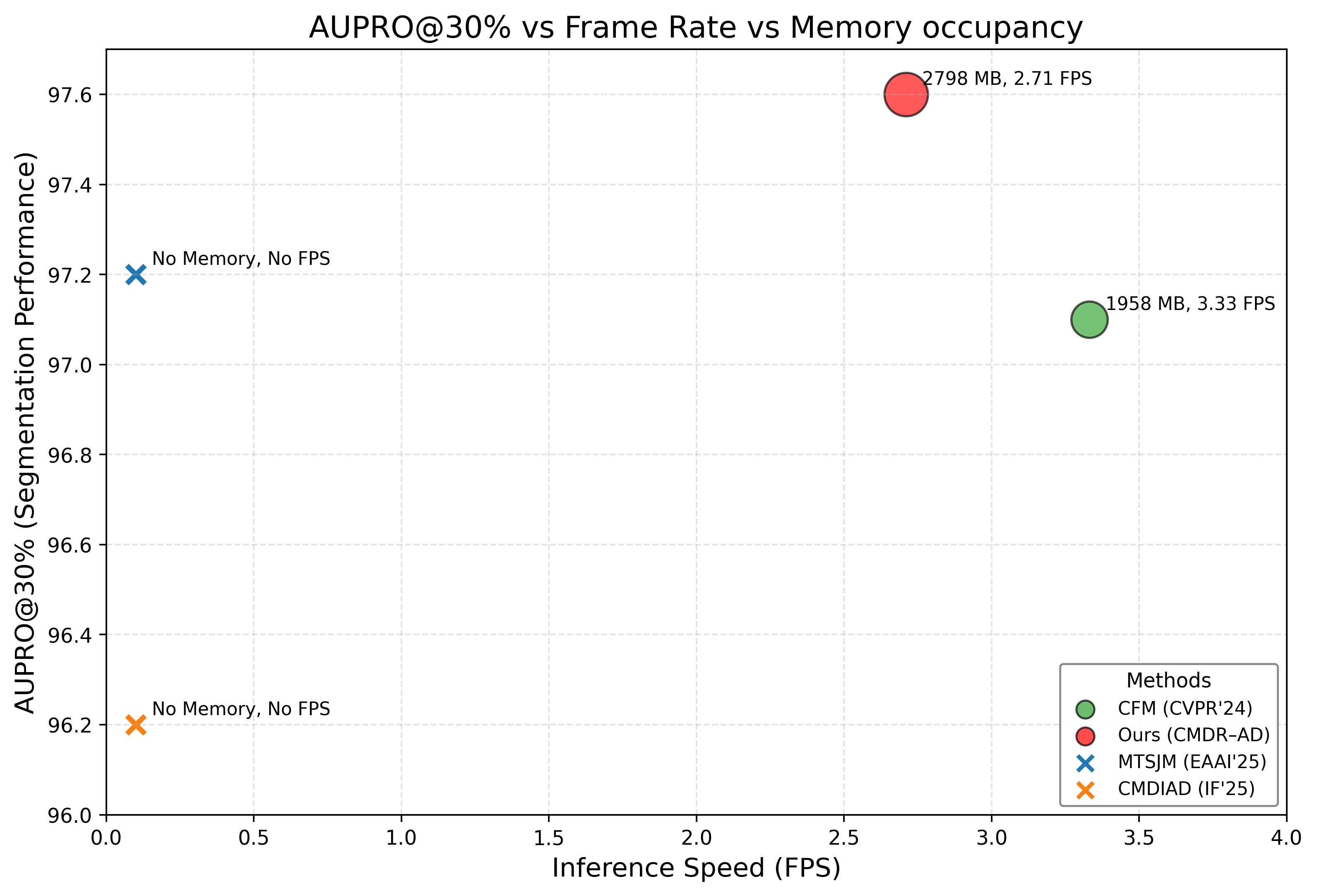} 
    \caption{\textbf{Performance, speed and memory occupancy of Multimodal Anomaly Detection methods.} The chart reports defect
segmentation performance (AUPRO@30\%) vs inference speed.}
    \label{fig:rec}
\end{figure}

\noindent The contributions presented in this paper can be summarized as follows:
\begin{itemize}
    \item We propose \textbf{CMDR--IAD}, a {modality-flexible} cross-modal mapping and dual-branch reconstruction framework for multimodal industrial anomaly detection, which explicitly models 2D--3D feature relationships while preserving modality-specific normal patterns.
    
    \item We introduce an adaptive fusion strategy that integrates cross-modal mapping inconsistencies with modality-aware reconstruction deviations through spatial reliability weighting and contrast-based gating, enabling robust anomaly localization under noisy or incomplete 3D measurements.
    
    \item We demonstrate that \textbf {CMDR--IAD} achieves state-of-the-art anomaly detection and localization performance on the MVTec 3D--AD benchmark while maintaining competitive inference speed and memory footprint (Fig.~\ref{fig:rec}), and we show strong performance of the 3D-only variant on a real polyurethane cutting dataset.
\end{itemize}

\section{Related Work}

\subsection{Unsupervised 2D Industrial Anomaly Detection}

Unsupervised 2D industrial anomaly detection methods \cite {liu2024deep} can be broadly grouped into reconstruction-based and feature-embedding-based approaches. Reconstruction-based models rely on autoencoders \cite{wang2020image, angiulli2023latent, bergmann2018improving, hou2021divide, ristea2022self, zavrtanik2021draem}, GANs \cite{ bougaham2024composite, yan2021learning}, inpainting networks \cite{ pirnay2022inpainting}, or diffusion models \cite{ wyatt2022anoddpm, teng2022unsupervised} to reproduce nominal samples and identify anomalies through reconstruction discrepancies. While intuitive, these methods often struggle due to their tendency to partially reconstruct abnormal regions.

\noindent Feature-embedding methods instead operate in the latent space \cite{li2021cutpaste, chiu2023self, gudovskiy2022cflow, salehi2021multiresolution, zhang2021anomaly} of deep networks.
Some approaches, such as Deep Feature Reconstruction (DFR) \cite{yang2020dfr}, train an autoencoder to reconstruct features extracted from normal samples and detect anomalies via feature-level reconstruction errors.
Another popular line of work relies on frozen feature extractors \cite{he2022masked, oquab2023dinov2, caron2021emerging, roth2022towards, bergman2020deep}, features from normal images are stored in a memory bank during training \cite{zou2022spot, lee2022cfa, roth2022towards}, and test features are compared against this bank to identify deviations.
These memory-based approaches achieve strong performance but often suffer from high memory usage and slow inference, as each test feature must be matched against many stored normal features.
\subsection{3D-Only Industrial Anomaly Detection}

Recent work in \emph{3D-only} anomaly detection focuses on identifying geometric defects directly from point clouds without relying on RGB appearance. Teacher--student methods, where a pretrained teacher models local geometric structure and a student learns to regress these features, have demonstrated strong capability in capturing subtle surface deviations~\cite{qin2023teacher, bergmann2023anomaly}. These approaches show that purely geometric information can be highly effective for detecting structural anomalies. This line of research is particularly relevant to the \emph{Polyurethane Cutting} dataset, where defects arise solely from geometric irregularities. In this setting, our 3D reconstruction pathway achieves strong performance, confirming that accurate modeling of surface geometry alone is sufficient for reliable defect identification and localization.

\subsection{Multimodal RGB--3D Industrial Anomaly Detection}

The release of the MVTec 3D-AD dataset \cite{bergmann2021mvtec} has significantly accelerated research in industrial anomaly detection involving both RGB appearance and 3D geometry. Initial multimodal baselines explored reconstruction-based strategies---including GANs, VAEs, and autoencoders---to jointly model depth and texture cues. Teacher--student paradigms were later introduced to improve robustness; for example, AST \cite{rudolph2023asymmetric} employs asymmetric networks to prevent students from overfitting anomalous patterns in RGB--depth pairs, while DAK-Net \cite{zhang2026dynamic} introduces dynamic background-guided asymmetric distillation for RGB--3D anomaly detection. Other works enhanced geometric reasoning by integrating handcrafted or learned 3D descriptors, as demonstrated by Horwitz and Hoshen \cite{horwitz2023back}.

\noindent Memory-bank--driven multimodal approaches extend successful 2D methods to 3D settings. BTF \cite{horwitz2023back} augments frozen 2D CNN features with handcrafted point-cloud descriptors and stores them in a feature bank for nearest-neighbor retrieval. M3DM \cite{wang2023multimodal} advances this idea by adopting frozen Transformer-based RGB and 3D backbones trained via large-scale self-supervision, and by introducing a learnable module to fuse 2D and 3D features before inserting them into a large multimodal memory bank. While effective, M3DM suffers from heavy memory usage and slow inference due to expensive feature comparisons.

\noindent Inspired by this line of work, CFM \cite{costanzino2024multimodal} replaces memory banks with lightweight cross-modal feature mapping networks that predict one modality from the other and compute anomaly scores directly from feature discrepancies. Further improvements include CMDIAD \cite{sui2025incomplete}, which employs cross-modal distillation to handle missing or incomplete modalities during inference. MTSJM \cite{liu2025multimodal} jointly optimizes multimodal representations through mask-based teacher--student learning and multimodal feature banks, achieving strong performance but still relying on large memory structures. More recently, 3D-ADNAS \cite{long2025revisiting} leverages neural architecture search to identify optimal RGB--3D fusion designs, showing that multimodal AD performance is sensitive to backbone and fusion choices.

\noindent In contrast to multimodal fusion strategies in M3DM (2023) \cite{wang2023multimodal}, CFM (2024) \cite{costanzino2024multimodal}, and recent fusion architectures such as MTSJM (2025) \cite{liu2025multimodal}, CMDIAD (2025) \cite{sui2025incomplete}, and 3D-ADNAS (2025) \cite{long2025revisiting}, DAK-Net (2026) \cite{zhang2026dynamic}, \textbf{CMDR--IAD} introduces explicit 2D$\leftrightarrow$3D cross-modal mapping and dedicated appearance--geometry reconstruction, leading to state-of-the-art results on MVTec 3D-AD.

\section{Methodology}
\label{sec:metho}

Our proposed unsupervised \textbf{CMDR--IAD} (Cross-Modal Mapping and Dual-Branch Reconstruction for 2D--3D Multimodal Industrial Anomaly Detection) integrates complementary RGB images $I^{2D}$, and point clouds $P^{3D}$ to detect both appearance- and geometry-related anomalies. \textbf{CMDR--IAD} aligns 2D and 3D feature spaces via cross-modal mapping networks while learning modality-specific reconstruction branches that capture normal patterns unique to each modality.

\begin{figure}[ht]
     \centering
    \includegraphics[width=1\textwidth]{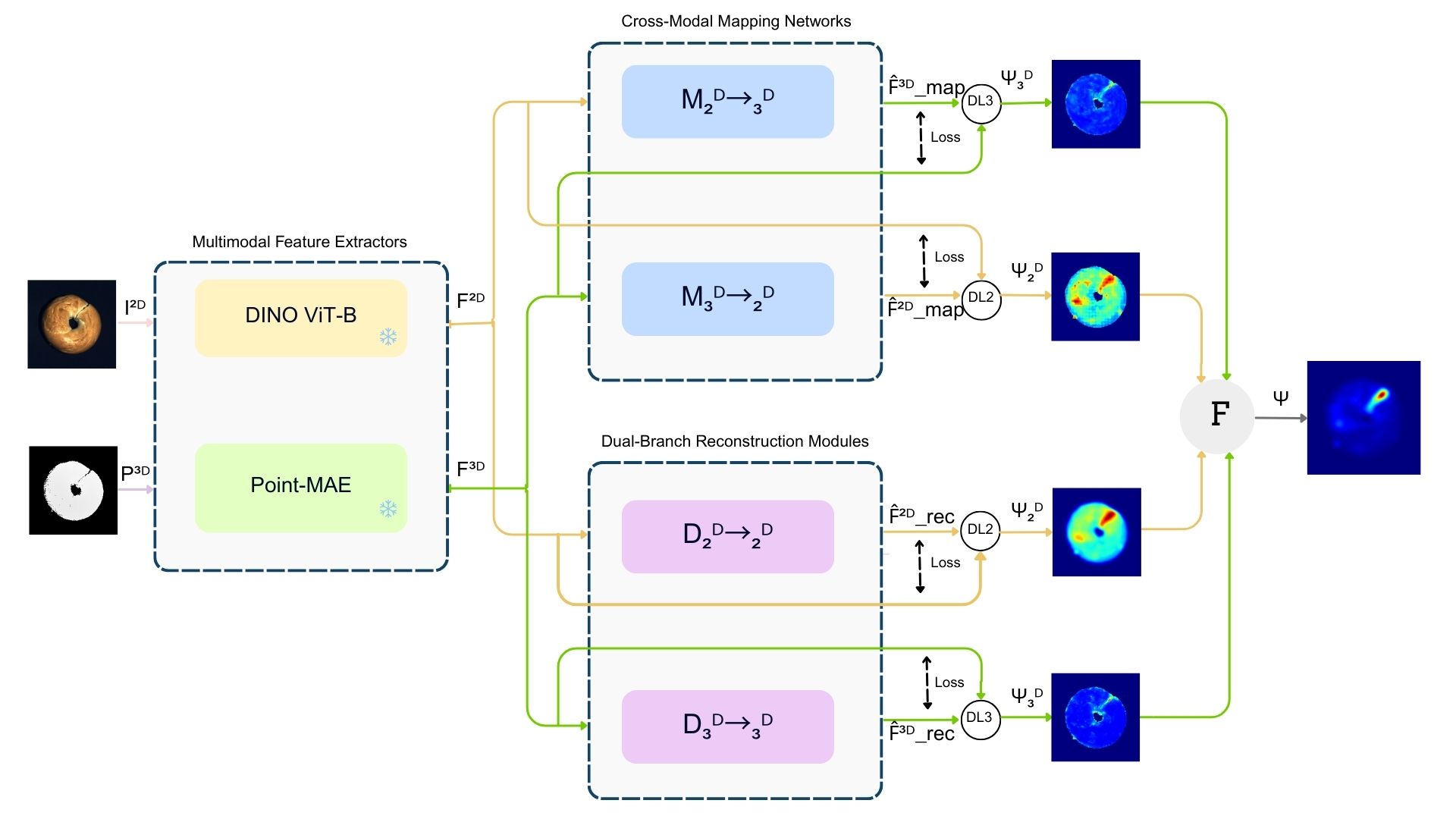} 
    \caption{\textbf{Overview of the proposed CMDR--IAD framework.} The method integrates RGB images and point clouds through multimodal feature extraction, cross-modal mapping, and dual-branch 2D--3D reconstruction. Cross-modal discrepancy maps and reconstruction errors are fused into the final anomaly map $\Psi$, enabling robust detection of both appearance- and geometry-related defects.}

    \label{fig:archi}
\end{figure}

\noindent As illustrated in Fig.~\ref{fig:archi}, the framework consists of four core components:
\begin{enumerate}
    \item \textbf{Multimodal Feature Extractors} -- frozen pretrained 2D and 3D encoders producing appearance features $F^{2D}$ and geometric features $F^{3D}$.
    \item \textbf{Cross-Modal Mapping Networks} -- lightweight MLPs projecting features between modalities to enforce 2D~$\leftrightarrow$~3D consistency.
    \item \textbf{Dual-Branch Reconstruction Modules} -- independent 2D and 3D decoders reconstructing modality-specific features to model normal appearance and geometry patterns.
   \item \textbf{Reliability-Aware Multimodal Fusion} -- a reliability- and confidence-aware fusion of cross-modal mapping discrepancies and reconstruction errors for robust anomaly localization.

   \end{enumerate}
\noindent\textbf{Modality-flexible operation.}
\textbf{CMDR--IAD} supports both multimodal and single-modality operation, enabling reliable anomaly detection with complete or missing sensory inputs.
Specifically, we consider two operational scenarios:
\begin{itemize}
    \item \textbf{Multimodal setting (MVTec 3D-AD)}: the full 2D+3D pipeline is employed, leveraging bidirectional cross-modal mapping together with both reconstruction branches to model appearance--geometry consistency.
    \item \textbf{3D-only setting (Polyurethane dataset)}: only the 3D pathway (Point-MAE encoder + Decoder3D) is used, supported by a dedicated point-cloud preprocessing pipeline.
\end{itemize}

\subsection{Multimodal Feature Extractors}

Appearance and geometric features are extracted from the RGB image $I^{2D}$ and point cloud $P^{3D}$ using pretrained encoders that remain frozen throughout training.

\noindent{\textbf{2D Feature Extraction.}}
Given an RGB image $I^{2D}$, the 2D encoder ${E}_{2D}$ (DINO ViT-B) produces a low-resolution feature map
$F^{2D}_{raw}$. Since its spatial resolution is smaller than that of the original image, we apply bilinear upsampling to obtain a dense pixel-aligned representation
\[
F^{2D} \in \mathbb{R}^{H \times W \times D_{2D}},
\]
providing one appearance feature vector per pixel.

\noindent{\textbf{3D Feature Extraction.}}
The point cloud $P^{3D}$ is processed by the 3D encoder ${E}_{3D}$ (Point-MAE/PointTransformer), producing a set of geometric feature tokens
$F^{3D}_{raw}$ associated with only a subset of points. Since many 3D feature extractors do not compute features for every input point but only for selected point groups \cite{ pang2023masked}, we interpolate features for all remaining points by using their nearest feature centers. Following a strategy similar to \cite{wang2023multimodal}, each point feature is computed from the features of its closest centers. This yields a dense per-point geometric representation
\[
F^{3D}_{pts} \in \mathbb{R}^{N \times D_{3D}}.
\]

\noindent{\textbf{Feature Alignment.}}
The pixel–point correspondences in MVTec 3D-AD \cite{bergmann2021mvtec} enable the integration of point-wise features into the image grid, yielding a dense representation of geometric information.
\[
F^{3D} \in \mathbb{R}^{H \times W \times D_{3D}}.
\]
Pixels without a corresponding 3D point are initialized to zero. To improve spatial coherence and suppress isolated artifacts introduced by sparse projections, we apply a light spatial smoothing implemented as a channel-wise 2D average pooling with a $3 \times 3$ kernel and stride $1$. Since no padding is used, the smoothed feature map is subsequently resized to the target resolution using adaptive average pooling.

\noindent At the end of this stage, the multimodal feature extractors produce two pixel-aligned feature maps:
\[
F^{2D} = {E}_{2D}(I^{2D}), \qquad
F^{3D} = {E}_{3D}(P^{3D}),
\]
which serve as the input to the cross-modal mapping networks and the dual-branch reconstruction modules.

\subsection{Cross-Modal Mapping Networks}

The cross-modal mapping stage explicitly models the correspondence between appearance features $F^{2D}$ and geometric features $F^{3D}$. \textbf{CMDR--IAD} employs two lightweight multilayer perceptrons (MLPs), $\mathcal{M}_{2D\rightarrow3D}$ and $\mathcal{M}_{3D\rightarrow2D}$, each composed of an input projection, a nonlinear transformation block (GELU + LayerNorm), and an output projection. For each pixel location $i$ with aligned multimodal features, the mappings predict the corresponding feature in the other modality as
\[
\hat{F}^{3D}_{map}(i) = \mathcal{M}_{2D\rightarrow3D}\big(F^{2D}(i)\big), 
\qquad
\hat{F}^{2D}_{map}(i) = \mathcal{M}_{3D\rightarrow2D}\big(F^{3D}(i)\big).
\]
If no valid 3D feature is available at location $i$ (e.g., due to missing depth or occlusion), the mapped feature is set to zero to avoid introducing invalid supervision. 
Setting mapped features to zero at locations with missing 3D observations prevents spurious cross-modal supervision from unreliable geometry, improving training stability. This masking avoids enforcing artificial appearance--geometry consistency in depth-sparse regions, thereby preserving anomaly sensitivity while allowing modality-specific reconstruction cues to dominate.

After processing all pixel locations, the mapping networks yield dense cross-modal prediction maps:
\[
\hat{F}^{3D}_{map} = \mathcal{M}_{2D\rightarrow3D}(F^{2D}), 
\qquad
\hat{F}^{2D}_{map} = \mathcal{M}_{3D\rightarrow2D}(F^{3D}).
\]

\subsection{Dual-Branch Reconstruction Modules}

To capture modality-specific normal patterns, \textbf{CMDR--IAD} employs two independent reconstruction branches: a 2D decoder $\mathcal{D}_{2D}$ and a 3D decoder $\mathcal{D}_{3D}$. Each branch operates solely within its own modality, enabling precise modeling of appearance cues in RGB images and geometric structures in point clouds.

\begin{figure}[h]
     \centering
    \includegraphics[width=1\textwidth]{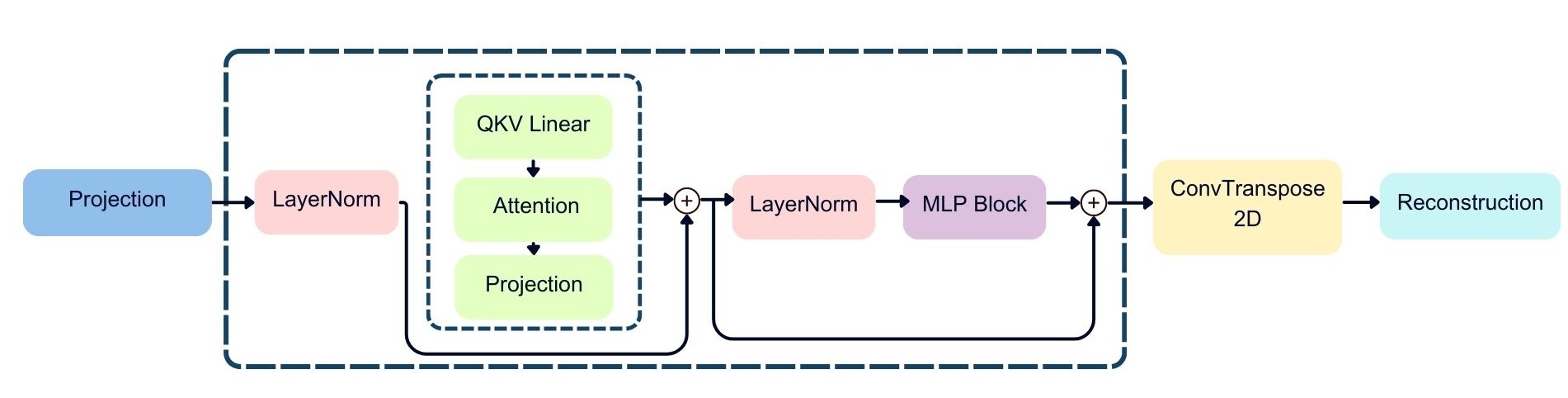} 
    \caption{\textbf{Overview of the 2D Reconstruction Branch.}
Projected features are enhanced using a sparse-attention block and an MLP refinement module with residual connections, then decoded through ConvTranspose2D layers to generate the reconstructed 2D feature map.}
    \label{fig:2d_recon}
\end{figure}

\noindent\textbf{2D Reconstruction Branch} (Fig.~\ref{fig:2d_recon}).
Given the 2D feature map $F^{2D}$, $\mathcal{D}_{2D}$ first projects the features to a compact latent space using linear layers with GELU and LayerNorm. A sparse-attention block is then applied to enhance local contextual dependencies, followed by an MLP refinement with residual connections. The refined features are reshaped into a spatial grid and decoded through ConvTranspose2D layers to produce the reconstructed feature map
\[
\hat{F}^{2D}_{rec} = \mathcal{D}_{2D}(F^{2D}).
\]

\begin{figure}[ht]
     \centering
    \includegraphics[width=1\textwidth]{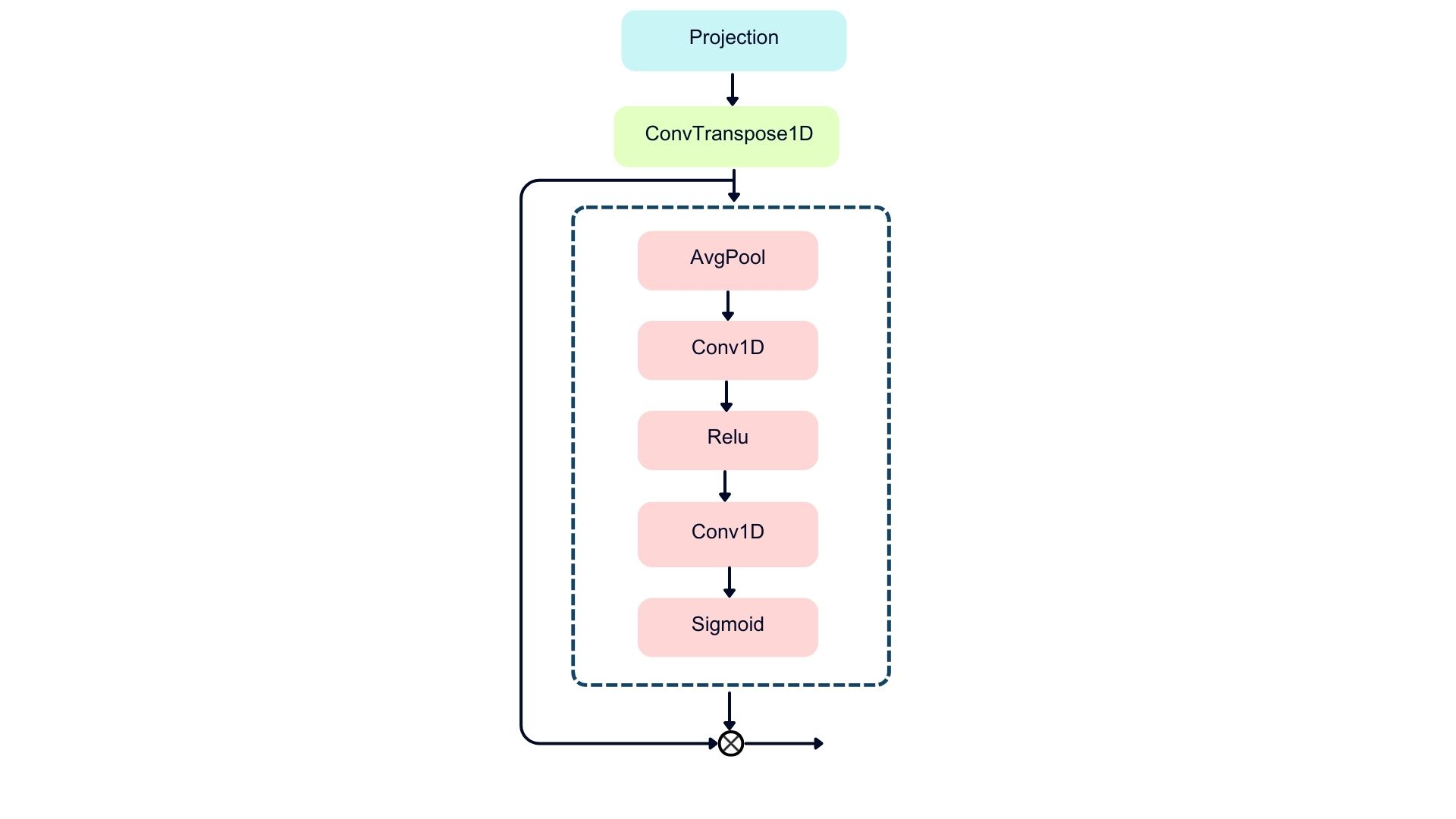} 
    \caption{\textbf{Overview of the 3D Reconstruction Branch.}
After projection and ConvTranspose1D upsampling, a gating block refines the features through sequential 1D convolutions and a sigmoid mask, which is multiplied with a residual pathway to produce the final 3D reconstruction.}
    \label{fig:3d_recon}
\end{figure}

\noindent\textbf{3D Reconstruction Branch} (Fig.~\ref{fig:3d_recon}).
The 3D decoder $\mathcal{D}_{3D}$ processes the geometric feature map $F^{3D}$ through a projection layer and a sequence of ConvTranspose1D layers to restore the original resolution. A lightweight channel-attention module is applied to emphasize informative geometric features, yielding the reconstructed representation
\[
\hat{F}^{3D}_{rec} = \mathcal{D}_{3D}(F^{3D}).
\]

\subsection{Training Objective}
\textbf{CMDR--IAD} optimizes the cross-modal mapping networks and reconstruction decoders independently, without a joint loss. Each module is trained using a masked cosine similarity loss computed only at locations with valid 2D--3D correspondences.

\noindent{\textbf{{Masked Similarity Loss.}}
We build upon the cosine similarity formulation commonly used in prior feature-alignment work \cite{costanzino2024multimodal} and adapt it to a masked setting.

\noindent Given a predicted feature $\hat{F}$ and its target feature $F$, we define
\[
\mathcal{L}_{sim}(F,\hat{F}) = 1 - \cos(F,\hat{F}),
\]
which is evaluated only over pixels with valid 3D features using a binary mask.

\noindent\textbf{Mapping and Reconstruction Losses.}
The two mapping networks are trained as
\[
\mathcal{L}_{map}^{2D} = \mathcal{L}_{sim}\!\left(F^{2D}, \hat{F}^{2D}_{map}\right), 
\qquad
\mathcal{L}_{map}^{3D} = \mathcal{L}_{sim}\!\left(F^{3D}, \hat{F}^{3D}_{map}\right),
\]
while the reconstruction decoders are optimized via
\[
\mathcal{L}_{rec}^{2D} = \mathcal{L}_{sim}\!\left(F^{2D}, \hat{F}^{2D}_{rec}\right), 
\qquad
\mathcal{L}_{rec}^{3D} = \mathcal{L}_{sim}\!\left(F^{3D}, \hat{F}^{3D}_{rec}\right).
\]

\noindent Each loss updates only its corresponding module, allowing all branches to be trained independently. This design avoids cross-modal gradient interference and promotes stable and specialized learning
\subsection{Anomaly Scoring and Reliability-Aware Multimodal Fusion}

 At inference time, \textbf{CMDR--IAD} computes anomaly scores by jointly analyzing cross-modal mapping discrepancies and modality-specific reconstruction errors. Feature discrepancies are measured using a normalized Euclidean distance. Given two feature vectors $a$ and $b$, we define
\[
d(a,b) = \left\lVert \frac{a}{\lVert a \rVert_2} - \frac{b}{\lVert b \rVert_2} \right\rVert_2 ,
\]
which corresponds to the $\ell_2$ distance between $\ell_2$-normalized features.

\noindent\textbf{Reliability-Gated Mapping Anomaly.}
Cross-modal mapping discrepancies are computed independently for each modality as
\[
d^{2D}_{map}(x,y) = d\!\left(\hat{F}^{2D}_{map}(x,y), F^{2D}(x,y)\right), \qquad
d^{3D}_{map}(x,y) = d\!\left(\hat{F}^{3D}_{map}(x,y), F^{3D}(x,y)\right).
\]
To emphasize spatially consistent appearance--geometry inconsistencies, the two discrepancies are combined into a joint mapping anomaly
\[
d_{\text{joint}}(x,y) = d^{2D}_{map}(x,y)\cdot d^{3D}_{map}(x,y).
\]
A spatial reliability gate $\alpha(x,y)\in[0,1]$ is derived from local statistics of $d_{\text{joint}}$ and reflects the confidence of cross-modal agreement. The resulting reliability-gated mapping anomaly is defined as
\[
A_{\text{map}}(x,y) = \alpha(x,y)\, d_{\text{joint}}(x,y).
\]

\noindent\textbf{Confidence-Weighted Reconstruction Anomaly.}
Reconstruction discrepancies are computed independently for each modality:
\[
d^{2D}_{rec}(x,y) = d\!\left(\hat{F}^{2D}_{rec}(x,y), F^{2D}(x,y)\right), \qquad
d^{3D}_{rec}(x,y) = d\!\left(\hat{F}^{3D}_{rec}(x,y), F^{3D}(x,y)\right).
\]
Each modality is assigned a confidence weight inversely related to its reconstruction error:
\[
w_2(x,y) = \exp\!\left(-B\, d^{2D}_{rec}(x,y)\right), \qquad
w_3(x,y) = \exp\!\left(-B\, d^{3D}_{rec}(x,y)\right),
\]
where $B$ is a temperature parameter controlling the sharpness of the weighting, set to $B=0.3$ in all experiments.
 
\noindent The confidence-weighted reconstruction anomaly is then defined as

\[
A_{\text{rec}}(x,y)
=
\frac{
w_2(x,y)\, d^{2D}_{rec}(x,y)
+
w_3(x,y)\, d^{3D}_{rec}(x,y)
}{
w_2(x,y) + w_3(x,y) + \varepsilon
}.
\]

\noindent\textbf{Final Anomaly Map.}
The final pixel-level anomaly score integrates cross-modal reliability and modality-specific confidence as
\[
\Psi_{\text{full}}(x,y)
=
A_{\text{map}}(x,y)\cdot A_{\text{rec}}(x,y),
\]
followed by light spatial smoothing using box filters to improve localization stability.

\noindent\textbf{Image-Level Anomaly Scoring.}
The smoothed anomaly map $\Psi$ is normalized over valid pixels. For image-level anomaly detection, we compute a scalar score
\[
S = \max_{x,y}
\frac{\Psi(x,y)}{\sqrt{\overline{\Psi}_{valid} + \varepsilon}},
\]
where $\overline{\Psi}_{valid}$ denotes the mean of $\Psi$ over unmasked pixels. The score $S$ is used for image-level anomaly detection, while $\Psi$ provides fine-grained pixel-level anomaly localization.

\subsection{3D-Only Inference Mode for Polyurethane Cuts}

The polyurethane cutting dataset provides only 3D point clouds without
registered RGB images. Accordingly, \textbf{CMDR--IAD} operates in a dedicated
\emph{3D-only inference mode}, where only the geometric branch of the framework
is activated. In this setting, raw point clouds are first processed using the
preprocessing pipeline described in in Sec.~\ref{sec:Preprocessing} and then passed to the frozen 3D
encoder ${E}_{3D}$ and the 3D decoder $\mathcal{D}_{3D}$ to identify structural
irregularities on cutting surfaces.

\noindent\textbf{Feature Extraction.}
Each point cloud $P^{3D}$ is processed by ${E}_{3D}$ to produce a dense
geometric feature representation:
\[
F^{3D} = {E}_{3D}(P^{3D}) .
\]
Since no RGB modality is available, cross-modal mapping and 2D reconstruction
are disabled.

\noindent\textbf{3D Reconstruction and Anomaly Scoring.}
The decoder $\mathcal{D}_{3D}$ reconstructs the expected normal geometry:
\[
\hat{F}^{3D}_{rec} = \mathcal{D}_{3D}(F^{3D}) .
\]
Per-point anomaly scores are computed as the normalized Euclidean distance
\[
A^{3D}_{rec}(i) =
\left\|
\tilde{F}^{3D}(i) - \tilde{\hat{F}}^{3D}_{rec}(i)
\right\|_2 ,
\]
where $\ell_2$-normalization is applied to both feature vectors. The resulting anomaly map is smoothed using the same box-filter strategy adopted in the multimodal setting.

\noindent\textbf{Image-Level Decision.}
An image-level anomaly score is obtained by max pooling:
\[
S = \max_{i} A^{3D}_{rec}(i),
\]
which is used for image-level AUROC evaluation, while $A^{3D}_{rec}$ supports
pixel-level AUROC computation.

\noindent
Despite the absence of RGB cues, the 3D-only variant of \textbf{CMDR--IAD} remains effective
for polyurethane cuts, as defects such as irregular cuts, burrs, gaps, and shape
distortions are strongly expressed in the geometric modality. The robust feature
representation of ${E}_{3D}$ enables reliable anomaly detection in this
unimodal setting.

\section{Datasets and Preprocessing}
\label{sec:Preprocessing}

To evaluate \textbf{CMDR--IAD} under both multimodal and single-modality conditions, we
use two datasets: MVTec 3D--AD for RGB+3D evaluation and a real-world
polyurethane cutting dataset to validate the 3D-only branch of the model.

\subsection{MVTec 3D--AD Setup}
MVTec 3D--AD contains 10 industrial object categories with 2656 training samples
and 1197 test samples \cite{bergmann2021mvtec}. Following the standard anomaly detection protocol, only
normal samples are used for training, while the test set includes both normal
and defective instances. Each sample provides an RGB image and a pixel-registered
3D representation storing $(x,y,z)$ coordinates.

\noindent MVTec 3D--AD serves as the primary multimodal benchmark for \textbf{CMDR--IAD}. In this
setting, the full 2D+3D pipeline is employed during inference, including both
feature extractors, cross-modal mapping networks, and reconstruction branches,
enabling evaluation of appearance--geometry consistency.

\subsection{Polyurethane 3D Dataset and Preprocessing Pipeline}

The polyurethane cutting dataset was collected within the \textbf{MOROSAI} project 
using a dual-sensor profilometer with a 405\,nm laser. The scanning system 
outputs dense 3D point clouds focused on the cutting edges of large 
polyurethane samples. Since only geometric data is available, this dataset 
is used to evaluate the 3D-only inference mode of \textbf{CMDR--IAD}.

\noindent The raw point clouds are processed by a three-stage preprocessing pipeline.

\noindent \textbf{Extraction of Geometric Outliers Using Isolation Forest}

\noindent Given a point cloud
\[
P = \{p_i \in \mathbb{R}^3\}_{i=1}^{N},
\]
we apply an unsupervised Isolation Forest (ISO) \cite{liu2008isolation} to detect sparse geometric
outliers caused by cutting defects or sensor noise. ISO is a tree-based
unsupervised anomaly detection method and is well suited for large-scale,
high-dimensional point cloud data. 
A low-contamination setting ($\gamma = 0.0001$) is adopted due to the sparsity of polyurethane defects, as
higher contamination levels may lead to false positives. ISO outputs a binary
point-level abnormality mask for each scan, used in subsequent processing.

\noindent \textbf{Sequential Chunking and Chunk-Level Labeling}

Each point cloud is sequentially partitioned into fixed-size chunks of
9216 points, following common practice for large-scale point cloud processing~\cite{hackel2017semantic3d}.

\noindent Let $C_k$ denote the $k$-th chunk and $m_i$ the ISO-based binary
outlier indicator for point $p_i$. The abnormal ratio of each chunk is
\[
f_k = \frac{1}{|C_k|} \sum_{p_i \in C_k} m_i.
\]
A chunk is labeled anomalous if $f_k \ge \tau$, with $\tau = 0.0025$.
For each scan, we store the chunked point clouds, their chunk-level labels,
and the corresponding per-point abnormal masks.

\noindent \textbf{Global Dataset Construction and Normalization}

All chunks are aggregated into global data, label, and point-level mask files.
Following a one-class anomaly detection protocol, only normal chunks are used
for training, with a 90/10 split applied to the normal subset. This results in
1856 normal samples for training and a test set of 235 samples, comprising
207 normal and 28 anomalous instances.

\noindent Each chunk is independently normalized using min--max scaling to the range
$[-1,1]$~\cite{lecun2012efficient}.
The resulting train and test splits, together with their corresponding
point-level masks are used to train and evaluate the 3D reconstruction branch
of \textbf{CMDR--IAD}.

\section{Experiments}

\subsection{Datasets and Evaluation Metrics}

We evaluate \textbf{CMDR--IAD} on the MVTec 3D--AD benchmark \cite{bergmann2021mvtec} and a real polyurethane
cutting dataset (Section~\ref{sec:Preprocessing}). MVTec 3D--AD provides
pixel-aligned RGB and 3D data for multimodal evaluation, while the polyurethane dataset consists of sequential 3D point-cloud blocks and is used to assess the 3D-only inference mode.

\noindent Performance is evaluated following the MVTec 3D--AD protocol. Image-level anomaly
detection performance is measured using the Area Under the Receiver Operating Characteristic Curve (I-AUROC). Pixel-level anomaly localization is primarily evaluated using the Area Under the Per-Region Overlap (AUPRO).
The experimental evaluation in the main paper is conducted using I-AUROC\%, AUPRO@30\% and AUPRO@1\%. Pixel-level AUROC (P-AUROC) results
are provided in the Appendix (Table~\ref{tab:pauroc_mvtec3dad}) for completeness.

\noindent For the polyurethane dataset, I-AUROC corresponds to block-level anomaly
classification, while point-level localization is evaluated using P-AUROC
against ISO-based ground-truth masks. This dual evaluation protocol enables
assessment of both coarse and fine-grained anomaly behavior in the 3D-only
setting.

\subsection{Implementation Details}

Following M3DM~\cite{wang2023multimodal} and CFM~\cite{costanzino2024multimodal},
we employ frozen pretrained Transformers as feature extractors: DINO ViT-B/8
\cite{caron2021emerging} trained on ImageNet~\cite{deng2009imagenet}
for RGB images, and Point-MAE~\cite{pang2023masked} trained on ShapeNet~\cite{chang2015shapenet}
for point clouds. For MVTec 3D--AD, both RGB images and pixel-aligned 3D features
are aligned on a $224\times224$ spatial grid .

\noindent Only the cross-modal mapping networks and the reconstruction decoders are trained.
For each MVTec 3D--AD category, the four modules
$\mathcal{M}_{2D\rightarrow3D}$, $\mathcal{M}_{3D\rightarrow2D}$,
$\mathcal{D}_{2D}$, and $\mathcal{D}_{3D}$ are optimized independently using the
masked cosine similarity loss described in Sec \ref{sec:metho}. Training is performed with
the Adam optimizer \cite{kingma2014adam}, learning rate $10^{-3}$, batch size~1, and 50 epochs.

\noindent For the polyurethane cutting dataset, \textbf{CMDR--IAD} operates in a 3D-only setting.
Only the 3D reconstruction decoder is trained using Point-MAE features. Each
sample is represented as a $96\times96$ pseudo-image corresponding to 9216
points, and training is performed with AdamW \cite{loshchilov2017decoupled}, learning rate $10^{-3}$, batch size~4, for 50 epochs.

\noindent Experimental results are obtained on a workstation configured with an NVIDIA A100 GPU (40960 MiB).

\noindent\textbf{Inference Speed and Memory Evaluation.}
Inference speed is measured in frames per second (FPS) on the same hardware for all datasets, including MVTec 3D--AD and the polyurethane cutting dataset.
We report the average inference time over all test samples, including input pre-processing and all subsequent steps up to anomaly score computation.
All GPU operations are synchronized before timing. Memory footprint is measured during inference and accounts for network parameters, intermediate activations, and auxiliary buffers.

\subsection{Results on MVTec 3D-AD dataset}
\subsubsection{Quantitative Results.}

Table~\ref{tab:mvtec3dad} reports the image-level I-AUROC results on the MVTec 3D-AD dataset. We compare our method with recent single-modality and multimodal approaches, including AST~\cite{rudolph2023asymmetric}, M3DM~\cite{wang2023multimodal}, CFM~\cite{costanzino2024multimodal}, CMDIAD~\cite{sui2025incomplete}, MTSJM~\cite{liu2025multimodal},
3D-ADNAS~\cite{long2025revisiting},
and DAK-Net~\cite{zhang2026dynamic}.

\noindent \textbf{2D results.}
In the 2D setting, \textbf{CMDR--IAD} achieves the highest I-AUROC on
\emph{Peach}, \emph{Potato}, and \emph{Tire}, being the only method exceeding 90\% on both \emph{Potato} and \emph{Tire}. It outperforms the second-best methods by more than 16\% on \emph{Potato} and 11.7\% on \emph{Tire}, while attaining the second-highest performance on \emph{Rope}. The mean I-AUROC of 87.5\% remains competitive with strong baselines such as AST, M3DM, MTSJM, and CMDIAD, despite 3D-ADNAS achieving the highest overall 2D mean.

\noindent \textbf{3D results.}
In the 3D setting, \textbf{CMDR--IAD} achieves the highest I-AUROC on
\emph{Cable Gland}, \emph{Foam}, \emph{Rope}, and \emph{Tire}, while attaining the
second-highest performance on \emph{Peach}. Notably, it is the only method
exceeding 90\% on \emph{Foam} and \emph{Tire}, outperforming the second-best
results by 14.6\% and 8.3\%, respectively. \textbf{CMDR--IAD} also achieves the
second-highest mean I-AUROC (87.5\%), demonstrating strong and consistent
performance across categories.

\noindent \textbf{2D+3D multimodal results.}  
In the multimodal setting, \textbf{CMDR--IAD} achieves the highest I-AUROC on 5 out of 10 object categories. Notably, \textbf{CMDR--IAD} is the only method whose I-AUROC consistently exceeds 93\% across all classes, with scores ranging from 93.0\% to 99.8\%. This leads to a mean I-AUROC of 97.3\%, surpassing strong multimodal baselines such as AST, M3DM, CFM, MTSJM, CMDIAD, 3D-ADNAS, and DAK-Net by up to 1.6\%, thereby establishing a new state of the art on MVTec 3D--AD. These results highlight the effectiveness of our reliability-aware fusion, which jointly leverages reconstruction confidence and cross-modal consistency to achieve robust and well-balanced anomaly detection across diverse object categories.

\begin{table*}[htbp]
\centering
\caption{I-AUROC (\%) for anomaly detection on all classes of the MVTec 3D-AD dataset. The best results are highlighted in \textbf{bold}, while the second-best are \underline{underlined}. Our approach delivers competitive performance in single-modality (2D and 3D) settings and achieves state-of-the-art results in the multimodal (2D+3D) configuration.}
\label{tab:mvtec3dad}
\resizebox{\textwidth}{!}{
\begin{tabular}{lccccccccccc}
\toprule
\textbf{Method} & \textbf{Bagel} & \textbf{Cable Gland} & \textbf{Carrot} & \textbf{Cookie} & \textbf{Dowel} & \textbf{Foam} & \textbf{Peach} & \textbf{Potato} & \textbf{Rope} & \textbf{Tire} & \textbf{Mean} \\
\midrule
\multicolumn{12}{c}{\textbf{2D}} \\
\midrule

PatchCore (2022) \cite{roth2022towards}
& 87.6 & 88.0 & 79.1 & 68.2 & 91.2 & 70.1 & 69.5 & 61.8 & 84.1 & 70.2 & 77.0 \\

BTF (2023)\cite{horwitz2023back}
& 87.6 & 88.0 & 79.1 & 68.2 & 91.2 & 70.1 & 69.5 & 61.8 & 84.1 & 70.2 & 77.0 \\

BTM (2024) \cite{lin2024back} & 90.9 & 89.5 & 83.8 & 74.5 & \underline{97.5} & 71.4 & 79.0 & 60.5 & 93.0 & 75.9 & 81.6 \\
AST (2023) \cite{rudolph2023asymmetric} & 94.7 & \underline{92.8} & 85.1 & \underline{82.5} & \textbf{98.1} &  \textbf{95.1} & 89.5 & 61.3 & \textbf{99.2} & 82.1 & \underline{88.0} \\
M3DM (2023) \cite{wang2023multimodal} & 94.4 & 91.8 & 89.6 & 74.9 & 95.9 & 76.7 & 91.9 & 64.8 & 93.8 & 76.7 & 85.0 \\
{MTSJM (2025) \cite{liu2025multimodal}} &\underline{97.1} &{90.8} & \textbf{94.2} & {73.1} &{96.7} & {79.5} & \underline{98.4} &{60.8} & {83.8} &{83.3} & {85.8} \\
CMDIAD (2025) \cite{sui2025incomplete} & 94.2 & 91.8 & 89.6 & 74.9 & 95.9 & 76.7 & 91.9 & 64.8 & 94.1 & 76.8 & 85.1 \\
3D-ADNAS (2025) \cite{long2025revisiting} & \textbf{98.1} & \textbf{98.8} & \underline{92.7} & \textbf{95.6} & 94.2 & \underline{92.8} & 85.3 & \underline{79.1} & 97.7 & \underline{85.8} &\textbf{92.0}\\


\textbf{{CMDR--IAD} (Ours)} &
82.4 & 63.6 & 77.8 & 81.0 & 95.4 & 84.3 & \textbf{99.0} & \textbf{95.2} & \underline{98.4} & \textbf{97.5} & 87.5\\

\midrule
\multicolumn{12}{c}{\textbf{3D}} \\
\midrule
Depth GAN (2022) \cite{bergmann2021mvtec}
& 53.0 & 37.6 & 60.7 & 60.3 & 49.7 & 48.4 & 59.5 & 48.9 & 53.6 & 52.1 & 52.4 \\

Depth AE (2022) \cite{bergmann2021mvtec}
& 46.8 & 73.1 & 49.7 & 67.3 & 53.4 & 41.7 & 48.5 & 54.9 & 56.4 & 54.6 & 54.6 \\

Depth VM (2022) \cite{bergmann2021mvtec}
& 51.0 & 54.2 & 46.9 & 57.6 & 60.9 & 69.9 & 45.0 & 41.9 & 66.8 & 52.0 & 54.6 \\

Voxel GAN (2022) \cite{bergmann2021mvtec}
& 38.3 & 62.3 & 47.4 & 63.9 & 56.4 & 40.9 & 61.7 & 42.7 & 66.3 & 57.7 & 53.8 \\

Voxel AE (2022) \cite{bergmann2021mvtec}
& 69.3 & 42.5 & 51.5 & 79.0 & 49.4 & 55.8 & 53.7 & 48.4 & 63.9 & 58.3 & 57.2 \\

Voxel VM (2022) \cite{bergmann2021mvtec}
& 75.0 & 74.7 & 61.3 & 73.8 & 82.3 & 69.3 & 67.9 & 65.2 & 60.9 & 69.0 & 69.9 \\

3D-ST (2023) \cite{bergmann2023anomaly}
& 86.2 & 48.4 & 83.2 & 89.4 & 84.8 & 66.3 & 76.3 & 68.7 & 95.8 & 48.6 & 74.8 \\

BTF (2023) \cite{horwitz2023back}
& 82.5 & 55.1 & 95.2 & 79.7 & \underline{88.3} & 58.2 & 75.8 & 88.9 & 92.9 & 65.3 & 78.2 \\
BTM (2024)\cite{lin2024back} & 93.9 & 55.3 & 91.6 & 84.4 & 82.3 & 58.8 & 71.8 & 92.8 & \underline{97.6} & 63.3 & 79.2 \\
AST (2023) \cite{rudolph2023asymmetric} & 88.1 & 57.6 & \textbf{96.5} & 95.7 & 67.9 & \underline{79.7} & \textbf{99.0} & 91.5 & 95.6 & 61.1 & 83.3 \\
M3DM (2023)  \cite{wang2023multimodal} & 94.1 & 65.1 & \textbf{96.5} & \underline{96.9} & \textbf{90.5} & 76.0 & 88.0 & \underline{97.4} & 92.6 & 76.5 & 87.4 \\
{MTSJM (2025) \cite{liu2025multimodal}} & \textbf{98.5} & {72.8} & \underline{96.1} & \textbf{99.1} &{87.6} & {77.3} & {90.6} & {93.5} & {93.2} &{85.6} & \textbf{89.4} \\
CMDIAD (2025) \cite{sui2025incomplete} & \underline{97.3} & 68.7 & 92.7 & 96.5 & 83.8 & 73.2 & 85.7 & \textbf{98.7} & 86.4 & \underline{86.3 }& 86.0 \\
3D-ADNAS (2025) \cite{long2025revisiting} & 79.4 & \underline{85.7} & 69.9 & 94.6 & 69.5 & 68.6 & 70.5 & 87.3 & 95.3 & 66.7 & 78.8 \\

\textbf{{CMDR--IAD} (Ours)} & 68.4 & \textbf{89.9} & 81.9 & 84.2 & 86.5 & \textbf{94.3} & \underline{93.4} & 84.3 & \textbf{98.1} & \textbf{94.4} & \underline{87.5} \\

\midrule
\multicolumn{12}{c}{\textbf{2D+3D}} \\
\midrule
Depth GAN (2022) \cite{bergmann2021mvtec}
& 53.8 & 37.2 & 58.0 & 60.3 & 43.0 & 53.4 & 64.2 & 60.1 & 44.3 & 57.7 & 53.2 \\

Depth AE (2022) \cite{bergmann2021mvtec}
& 64.8 & 50.2 & 65.0 & 48.8 & 80.5 & 52.2 & 71.2 & 52.9 & 54.0 & 55.2 & 59.5 \\

Depth VM (2022) \cite{bergmann2021mvtec}
& 51.3 & 55.1 & 47.7 & 58.1 & 61.7 & 71.6 & 45.0 & 42.1 & 59.8 & 62.3 & 55.5 \\

Voxel GAN (2022) \cite{bergmann2021mvtec}
& 68.0 & 32.4 & 56.5 & 39.9 & 49.7 & 48.2 & 56.6 & 57.9 & 60.1 & 48.2 & 51.7 \\

Voxel AE (2022) \cite{bergmann2021mvtec}
& 51.0 & 54.0 & 38.4 & 69.3 & 44.6 & 63.2 & 55.0 & 49.4 & 72.1 & 41.3 & 53.8 \\

Voxel VM (2022) \cite{bergmann2021mvtec}
& 55.3 & 77.2 & 48.4 & 70.1 & 75.1 & 57.8 & 48.0 & 46.6 & 68.9 & 61.1 & 60.9 \\

3D-ST (2023) \cite{bergmann2023anomaly}
& 95.0 & 48.3 & \textbf{98.6} & 92.1 & 90.5 & 63.2 & 94.5 & \textbf{98.8} & 97.6 & 54.2 & 83.3 \\

BTM (2024) \cite{lin2024back} & 98.0 & 86.0 & 98.0 & 96.3 &  {97.8} & 72.6 & 95.8 & 95.3 &{ 98.0} &{ 92.6} & 93.0 \\
BTF (2023) \cite{horwitz2023back} & 91.8 & 74.8 & 96.7 & 88.3 & 93.2 & 58.2 & 89.6 & 91.2 & 92.1 & 88.6 & 86.5 \\

AST (2023) \cite{rudolph2023asymmetric} & 98.3 & 87.3 & 97.6 & 97.1 & 93.2 & 88.5 & 97.4 &\underline{ 98.1 }&\textbf {100.0} & 79.7 & 93.7 \\
M3DM (2023) \cite{wang2023multimodal} & 99.4 &  {90.9 }& 97.2 & 97.6 & 96.0 &  \underline{94.2} & 97.3 & 89.9 & 97.2 & 85.0 & 94.5 \\
CFM (2024) \cite{costanzino2024multimodal} & 99.4 & 88.8 & 98.4 &  {99.3} & {98.0} & 88.8 & 94.1 & 94.3 & 98.0 &  \underline{95.3} & 95.4 \\

{MTSJM (2025) \cite{liu2025multimodal}} & \textbf{100.0} &  \underline{93.1} & \underline{98.5} & \underline{99.4} & {96.8} & 89.9 &\underline{98.6} & 94.7 &96.2 & 89.7 & \underline{95.7} \\
CMDIAD (2025) \cite{sui2025incomplete} & 99.2 & 89.3 & 97.7 & 96.0 & 95.3 & 88.3 & 95.0 & 93.7 & 94.3 & 89.3 & 93.8 \\
3D-ADNAS (2025) \cite{long2025revisiting} &  \underline{99.7} & \textbf{100.0 }& 97.1 &  {98.6} & 96.6 &  \textbf{94.8 }& 89.7 & 87.3 & \textbf{100.0} & 86.7 & 95.1 \\
DAK-Net (2026) \cite{zhang2026dynamic} 
& 96.8 & 90.0 & 97.3 & 97.2 & \underline{98.6} & 93.9 & 95.2 & 87.6 & \underline{99.4} & 80.0 & 93.6 \\

\textbf{{CMDR--IAD} (Ours)}   & 99.6 & 93.0 & \textbf{98.6} & \textbf{99.8} &  \textbf{99.1} & 93.6 & \textbf{99.6} & 93.1 & {98.7} &  \textbf{97.5} & \textbf{97.3} \\


\bottomrule
\end{tabular}}
\end{table*}

\begin{table*}[htbp]
\centering
\caption{AUPRO@30\% for anomaly detection on all classes of the MVTec 3D-AD dataset. The best results are highlighted in \textbf{bold}, while the second-best are \underline{underlined}. Our approach delivers competitive performance in single-modality (2D and 3D) settings and achieves state-of-the-art localization accuracy in the multimodal (2D+3D) configuration.}

\label{tab:mvtec3dad_aupro}
\resizebox{\textwidth}{!}{
\begin{tabular}{lccccccccccc}
\toprule
\textbf{Method} & \textbf{Bagel} & \textbf{Cable Gland} & \textbf{Carrot} & \textbf{Cookie} & \textbf{Dowel} & \textbf{Foam} & \textbf{Peach} & \textbf{Potato} & \textbf{Rope} & \textbf{Tire} & \textbf{Mean} \\
\midrule
\multicolumn{12}{c}{\textbf{2D}} \\
\midrule

PatchCore (2022) \cite{roth2022towards}
& 90.1 & 94.9 & 92.8 & 87.7 & 89.2 & 56.3 & 90.4 & 93.2 & 90.8 & 90.6 & 87.6 \\

BTF (2023) \cite{horwitz2023back}
& 90.1 & 94.9 & 92.8 & 87.7 & 89.2 & 56.3 & 90.4 & 93.2 & 90.8 & 90.6 & 87.6 \\
BTM (2024) \cite{lin2024back} & 90.1 & 95.8 & 94.2 & 90.5 & \underline{95.1} & 61.5 & 90.6 & 93.8 & 92.7 & 91.6 & 89.6 \\
AST (2023) \cite{rudolph2023asymmetric} & 83.4 & 87.4 & 76.4 & 44.7 & 87.2 & 56.9 & 84.5 & 51.8 & 86.3 & 58.8 & 71.7 \\
M3DM (2023)  \cite{wang2023multimodal} & \underline{95.2} & \textbf{97.2} & \underline{97.3} & 89.1 & 93.2 & 84.3 & \underline{97.0} & 95.6 & \underline{96.8} & \textbf{96.6} &\underline{94.2} \\

{MTSJM (2025) \cite{liu2025multimodal}} & \textbf{97.6} &\underline{96.9} & \textbf{97.5} & \underline{91.4} & {94.9} & \textbf{86.9} & \textbf{97.5} & \underline{96.3} & \textbf{97.0} & \textbf{96.6} & \textbf{95.3} \\
CMDIAD (2025) \cite{sui2025incomplete}
& 95.1 & \textbf{97.2} & \underline{97.3} & 89.1 & 93.2 & 84.3 & \underline{97.0} & 95.6 & \underline{96.8} & \textbf{96.6} & \underline{94.2} \\

\textbf{{CMDR--IAD} (Ours)} & 
\underline{95.2} & 96.3 & 83.1 & \textbf{96.9} &\textbf{97.5} & \underline{85.1} & 90.7 & \textbf{98.1} & 96.3 & \underline{96.0} & 93.5 \\

\midrule
\multicolumn{12}{c}{\textbf{3D}} \\
\midrule
Depth GAN (2022) \cite{bergmann2021mvtec}
& 11.1 & 7.2 & 21.2 & 17.4 & 16.0 & 12.8 & 0.3 & 4.2 & 44.6 & 7.5 & 14.3 \\

Depth AE (2022) \cite{bergmann2021mvtec}
& 14.7 & 6.9 & 29.3 & 21.7 & 20.7 & 18.1 & 16.4 & 6.6 & 54.5 & 14.2 & 20.3 \\

Depth VM (2022) \cite{bergmann2021mvtec}
& 28.0 & 37.4 & 24.3 & 52.6 & 48.5 & 31.4 & 19.9 & 38.8 & 54.3 & 38.5 & 37.4 \\

Voxel GAN (2022) \cite{bergmann2021mvtec}
& 44.0 & 45.3 & 87.5 & 75.5 & 78.2 & 37.8 & 39.2 & 63.9 & 77.5 & 38.9 & 58.3 \\

Voxel AE (2022) \cite{bergmann2021mvtec}
& 26.0 & 34.1 & 58.1 & 35.1 & 50.2 & 23.4 & 35.1 & 65.8 & 1.5 & 18.5 & 34.8 \\

Voxel VM (2022) \cite{bergmann2021mvtec}
& 45.3 & 34.3 & 52.1 & 69.7 & 68.0 & 28.4 & 34.9 & 63.4 & 61.6 & 34.6 & 49.2 \\

BTF (2023) \cite{horwitz2023back} 
& \underline{97.3} & \underline{87.9} & \textbf{98.2} & 90.6 & 89.2 & 73.5 & \underline{97.7} & \underline{98.2} & 95.6 & \underline{96.1} & 92.4 \\

BTM (2024) \cite{lin2024back} & \textbf{97.4} & 86.1 & \underline{98.1} & \underline{93.7} & \underline{95.9} & 66.1 & \textbf{97.8} & \textbf{98.3} & \textbf{98.0} & 94.7 & \textbf{92.6} \\
AST (2023) \cite{rudolph2023asymmetric} & {95.4} & 72.3 & 97.4 & 90.1 & 83.2 & \underline{82.6} & 97.6 & \underline{98.2} & 89.2 & 75.1 & 88.1 \\
M3DM (2023)  \cite{wang2023multimodal} & 94.3 & 81.8 & 97.7 & 88.2 & 88.1 & 74.3 & 95.8 & 97.4 & 95.0 & 92.9 & 90.6 \\
{MTSJM (2025) \cite{liu2025multimodal}} & {92.9} & {85.5} & {97.1} &{88.7} & {87.8} & {76.4} &{94.0} & {97.1} &{93.7} &{95.1} &{90.8} \\

CMDIAD (2025) \cite{sui2025incomplete} & 94.7 & 82.6 & 97.7 & 88.2 & 88.1 & 76.7 & 96.7 & 97.8 & 94.7 & 94.0 & 91.1 \\

\textbf{{CMDR--IAD} (Ours)}& 81.2 & \textbf{97.9} & 76.3 & \textbf{96.6} & \textbf{97.9} & \textbf{92.8} & 90.8 & 97.4 & \underline{97.0} & \textbf{96.8} & \underline{92.5 }\\

\midrule
\multicolumn{12}{c}{\textbf{2D+3D}} \\
\midrule
Depth GAN (2022) \cite{bergmann2021mvtec}
& 42.1 & 42.2 & 77.8 & 69.6 & 49.4 & 25.2 & 28.5 & 36.2 & 40.2 & 63.1 & 47.4 \\

Depth AE ( 2022) \cite{bergmann2021mvtec}
& 43.2 & 15.8 & 80.8 & 49.1 & 84.1 & 40.6 & 26.2 & 21.6 & 71.6 & 47.8 & 48.1 \\

Depth VM (2022) \cite{bergmann2021mvtec}
& 38.8 & 32.1 & 19.4 & 57.0 & 40.8 & 28.2 & 24.4 & 34.9 & 26.8 & 33.1 & 33.5 \\

Voxel GAN (2022) \cite{bergmann2021mvtec}
& 66.4 & 62.0 & 76.6 & 74.0 & 78.3 & 33.2 & 58.2 & 79.0 & 63.3 & 48.3 & 63.9 \\

Voxel AE (2022) \cite{bergmann2021mvtec}
& 46.7 & 75.0 & 80.8 & 55.0 & 76.5 & 47.3 & 72.1 & 91.8 & 1.9 & 17.0 & 56.4 \\

Voxel VM (2022) \cite{bergmann2021mvtec}
& 51.0 & 33.1 & 41.3 & 71.5 & 68.0 & 27.9 & 30.0 & 50.7 & 61.1 & 36.6 & 47.1 \\

3D-ST ( 2023)\cite{bergmann2023anomaly}
& 95.0 & 48.3 &  \textbf{98.6} & 92.1 & 90.5 & 63.2 & 94.5 &  \textbf{98.8} & \underline{97.6} & 54.2 & 83.3 \\

BTF (2023) \cite{horwitz2023back}
& 97.6 & 96.9 & 97.9 & \underline{97.3 }& 93.3 & 88.8 & 97.5 & 98.1 & 95.0 & 97.1 & 95.9 \\

BTM (2024) \cite{lin2024back} & {97.9} & {97.2}& 98.0 &  \textbf{97.6} &  \textbf{97.7} & 90.5 & 97.8 & 98.2 & 96.8 & {97.5} & 96.9 \\
BTF (2023) \cite{horwitz2023back} & 97.6 & 96.9 & 97.9 & \underline{97.3} & 93.3 & 88.8 & 97.5 & 98.1 & 95.0 & 97.1 & 95.9 \\

AST (2023) \cite{rudolph2023asymmetric} & 97.0 & 94.7 & 98.1 & 93.6 & 91.2 & 91.8 & {97.9} & \underline{98.3} & 88.7 & 94.2 & 94.6 \\
M3DM (2023)  \cite{wang2023multimodal} & 97.0 & 97.1 & 97.9 & 95.0 & 94.1 & 93.2 & 97.7 & 97.1 & 97.1 & 97.5 & 96.4 \\
CFM(2024) \cite{costanzino2024multimodal} & 97.9 &  {97.2} & 98.2 & 94.5 & 95.0 & \underline{96.8 }&  \underline{98.0} & 98.2 & 97.5 &  \textbf{98.1} & 97.1 \\

{MTSJM (2025) \cite{liu2025multimodal}} & \textbf{98.4} &  \textbf{98.1} & \underline{98.3} & {96.8} & {93.9} & {95.1} &{97.7} & {98.5} &{97.4} & {97.3} & \underline{97.2} \\
CMDIAD (2025) \cite{sui2025incomplete} & 97.0 & 97.1 & 97.7 & 93.2 & 93.4 & 94.6 & 97.8 & 97.0 & 97.0 & 97.4 & 96.2 \\



\textbf{{CMDR--IAD} (Ours)} &  \underline{98.0} &  \underline{97.3} & 98.2 & 95.9 &  \underline{96.6} & \textbf{97.1} & \textbf{98.2} & 98.2 &\textbf{ 98.0} & \textbf{98.2} & \textbf{97.6} \\

\bottomrule
\end{tabular}}
\end{table*}

\noindent Table~\ref{tab:mvtec3dad_aupro} reports the AUPRO@30\% performance across 2D, 3D, and 2D+3D modalities on the MVTec 3D-AD dataset. AUPRO provides a strict and fine-grained evaluation of anomaly localization quality.

\noindent \textbf{2D results.} \textbf{CMDR--IAD} achieves a competitive mean AUPRO@30\%, clearly outperforming
reconstruction-based baselines such as AST and approaching strong
memory-based methods including M3DM and MTSJM. Our method attains the
highest AUPRO@30\% on \emph{Cookie}, \emph{Dowel}, and \emph{Potato}, and the
second-highest performance on \emph{Bagel}, \emph{Foam}, and \emph{Tire}.
These results indicate that the 2D branch effectively captures
discriminative texture cues and provides strong localization performance.

\noindent \textbf{3D results.} In the 3D setting, \textbf{CMDR--IAD} achieves the highest localization performance on
\emph{Cable Gland}, \emph{Cookie}, \emph{Dowel}, \emph{Foam}, and \emph{Tire},
while attaining the second-highest result on \emph{Rope}. Notably, it is the
only method exceeding 90\% AUPRO@30\% on \emph{Cable Gland} and \emph{Foam},
outperforming the second-best method by approximately 10\% and 10.2\%,
respectively. Overall, \textbf{CMDR--IAD} reaches a mean AUPRO@30\% of 92.5\%, which is
effectively identical to the highest-performing method, BTM (92.6\%), and
demonstrates the effectiveness of our point-cloud reconstruction in modeling
geometric structure and detecting subtle shape deviations.

\noindent \textbf{2D+3D multimodal results.} In the multimodal setting, \textbf{CMDR--IAD} achieves the highest AUPRO@30\% on four
object categories and the second-highest on three others. Notably, it is the
only method whose AUPRO@30\% consistently exceeds 95.5\% across all classes,
with scores ranging from 95.9\% to 98.2\%. This results in the highest mean
AUPRO@30\% of 97.6\%, outperforming all competing multimodal baselines. These
results underline the effectiveness of the proposed fusion strategy, which
integrates geometric reconstruction reliability with cross-modal consistency
for robust and well-balanced anomaly localization.

\begin{table}[ht]
\centering
\caption{AUPRO@1\% on the MVTec 3D-AD dataset for multimodal (2D+3D) anomaly detection. Best results are shown in \textbf{bold} and second-best results are \underline{underlined}.}

\label{tab:comparison_percent}

\resizebox{\linewidth}{!}{
\begin{tabular}{lccccccccccc}
\hline
Method & Bagel & CableGland & Carrot & Cookie & Dowel & Foam & Peach & Potato & Rope & Tire & Mean \\
\hline
BTF (2023) \cite{horwitz2023back}   & 42.8 & 36.5 & 45.2 & 43.1 & 37.0 & 24.4 & 42.7 & {47.0} & 29.8 & 34.5 & 38.3 \\
AST (2023) \cite{rudolph2023asymmetric}   & 38.8 & 32.2 & {47.0} & 41.1 & 32.8 & 27.5 &  \underline{47.4} & \textbf{48.7} & 36.0 & {47.4} & 39.8 \\
M3DM (2023) \cite{wang2023multimodal}  & 41.4 & 39.5 & 44.7 & 31.8 & \textbf{42.2} & 33.5 & 44.4 & 35.1 & 41.6 & 39.8 & 39.4 \\
CFM(2024) \cite{costanzino2024multimodal}   & \underline{45.9} & \underline{43.1} & \underline{48.5} & \textbf{46.9} & \underline{39.4} & \underline{41.3} & {46.8} & \textbf{48.7} & \underline{46.4 }& \underline{47.6} & \underline{45.5} \\

\textbf{{CMDR--IAD} (Ours)}        & \textbf{47.2} & \textbf{43.9} & \textbf{48.6} & \underline{46.4} & \textbf{42.2} &\textbf{ 43.5} & \textbf{48.6} & \underline{48.5} & \textbf{47.4} & \textbf{48.5} & \textbf{46.5} \\

\hline
\end{tabular}
}
\end{table}
\noindent Table~\ref{tab:comparison_percent} reports the AUPRO@1\% results for the
2D+3D multimodal setting on the MVTec~3D-AD dataset, which provides a strict
evaluation of anomaly localization under low false-positive constraints.
\textbf{CMDR--IAD} achieves the highest AUPRO@1\% on 8 out of 10 object categories and
the second-highest performance on the remaining two. Notably, it is the only
method whose AUPRO@1\% consistently exceeds 42 across all classes, with scores
ranging from 42.2 to 48.6. This leads to the highest mean AUPRO@1\% of 46.5\%,
demonstrating the robustness and effectiveness of our multimodal fusion in
capturing subtle appearance and geometric anomalies.
\begin{table}[ht]
\centering
\caption{Inference speed, memory footprint, and anomaly detection performance on MVTec 3D-AD. 
Frame Rate in FPS and Memory in MB.}
\label{tab:multimodal_efficiency}

\resizebox{\linewidth}{!}{
\begin{tabular}{lccccccc}
\hline
Method & Ref & Frame Rate (FPS) & Memory (MB) & I-AUROC & P-AUROC & AUPRO@30\% & AUPRO@1\% \\
\hline

CFM \cite{costanzino2024multimodal}       & CVPR 2024            & \textbf{3.331} & \textbf{1957.77}   & 95.4 & \underline{99.3} & 97.1 & \underline{45.5} \\
3D-ADNAS \cite{long2025revisiting}  & AAAI 2025            & -     & -      & 95.1 & -   & -   & - \\
MTSJM \cite{liu2025multimodal}    & EAAI 2025            & -     & -       & \underline{95.7} & - & \underline{97.2} & - \\
CMDIAD \cite{sui2025incomplete}    & Information Fusion 2025 & -  & -       & 93.8 & 99.2 & 96.2 & - \\
DAK-Net \cite{zhang2026dynamic}  & Journal of Manufacturing Processes 2026 & - & - & 93.6 & - & - & - \\

\textbf{{CMDR--IAD} (Ours)} & - & \underline{2.710} & \underline{2797.65}  & \textbf{97.3} & \textbf{99.6} & \textbf{97.6} & \textbf{46.5} \\

\hline
\end{tabular}
}
\end{table}

\noindent Table~\ref{tab:multimodal_efficiency} compares inference speed, memory footprint, and anomaly detection accuracy for representative multimodal methods on
MVTec~3D-AD. While CFM achieves the highest frame rate and lowest memory usage,
\textbf{CMDR--IAD} attains the best overall detection and localization performance,
achieving the highest I-AUROC, P-AUROC, AUPRO@30\%, and AUPRO@1\%. Despite a
moderate increase in memory consumption, our method maintains competitive
inference speed while significantly improving localization accuracy under both
standard and strict false-positive constraints. These results highlight that
\textbf{CMDR--IAD} offers a favorable balance between computational efficiency and
state-of-the-art anomaly detection performance.
\subsubsection{Qualitative Visualization}

Figs.~\ref{fig:Anomaly_map_mvtec_figure2d}--\ref{fig:Anomaly_map_mvtec_figure2d+3d} present qualitative anomaly localization results on representative MVTec 3D-AD categories under three settings: 2D-only, 3D-only, and multimodal (2D+3D). In all figures, warmer colors correspond to higher anomaly scores, while cooler colors indicate normal regions.

\noindent\textbf{2D-only results.}
As shown in Fig.~\ref{fig:Anomaly_map_mvtec_figure2d}, AST (2D) often suffers from under-detection, highlighting only limited abnormal regions, while M3DM (2D) tends to over-detect by assigning high scores to surrounding normal areas. MTSJM (2D) reduces false positives but still exhibits imprecise boundaries on complex defects. In contrast, \textbf{CMDR--IAD} (2D) produces more compact and accurate anomaly maps, effectively suppressing background noise.

\noindent \textbf{3D-only results.}
Fig.~\ref{fig:Anomaly_map_mvtec_figure3d} shows that AST (3D) frequently misses small or thin geometric defects, whereas M3DM (3D) overestimates anomalous regions on complex surfaces. MTSJM (3D) achieves improved localization but still presents scattered responses in some cases. By comparison, \textbf{CMDR--IAD} (3D) consistently highlights structurally abnormal regions with clearer boundaries, demonstrating robust 3D-only inference.
\begin{figure}[H]
    \centering
    \includegraphics[width=1\textwidth]{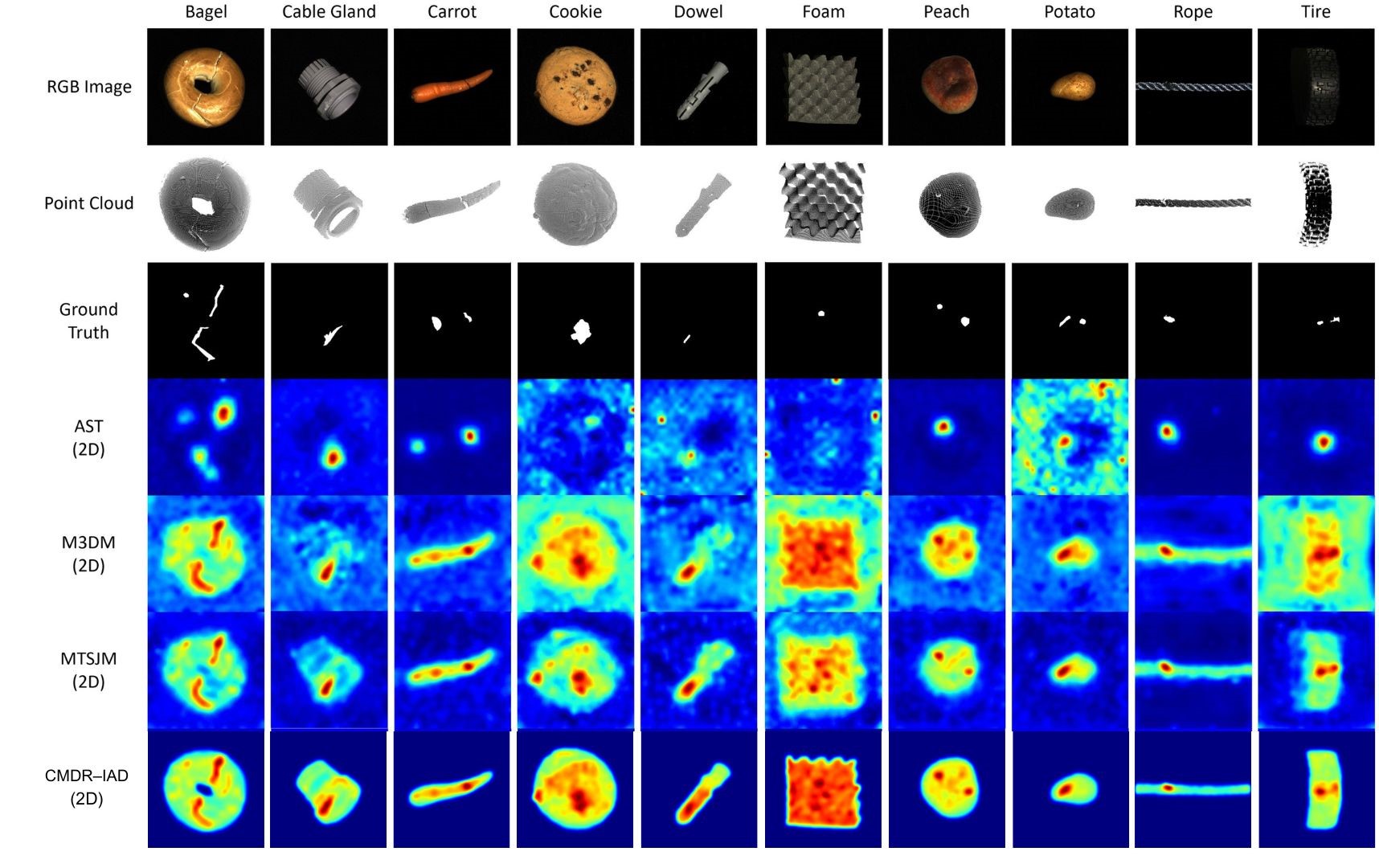}
    \caption{
Qualitative anomaly localization results on the MVTec 3D-AD dataset using \textbf{2D-only} features. From top to bottom, the rows show the RGB image, point cloud, ground truth, AST (2D), M3DM (2D), MTSJM (2D), and the proposed \textbf{CMDR-IAD} (2D). Warmer colors indicate higher anomaly scores, while cooler colors denote normal regions.
}
    \label{fig:Anomaly_map_mvtec_figure2d}
\end{figure}
\noindent \textbf{Multimodal results.}
As illustrated in Fig.~\ref{fig:Anomaly_map_mvtec_figure2d+3d}, combining RGB and 3D cues improves anomaly localization for all methods. However, AST and M3DM remain limited by weak cross-modal interaction and noisy feature differences, respectively. Benefiting from explicit cross-modal mapping and reliability-aware fusion, \textbf{CMDR--IAD} (2D+3D) achieves the most accurate and concentrated anomaly localization, closely matching the ground truth.

\noindent Overall, the qualitative results confirm that \textbf{CMDR--IAD} effectively mitigates the under-detection of AST and the over-detection of M3DM, while surpassing MTSJM in localization precision. These observations are consistent with the I-AUROC\% and AUPRO30\% improvements reported in Tables~\ref{tab:mvtec3dad} and~\ref{tab:mvtec3dad_aupro}.

\begin{figure}[!ht]
    \centering
    \includegraphics[width=1\textwidth]{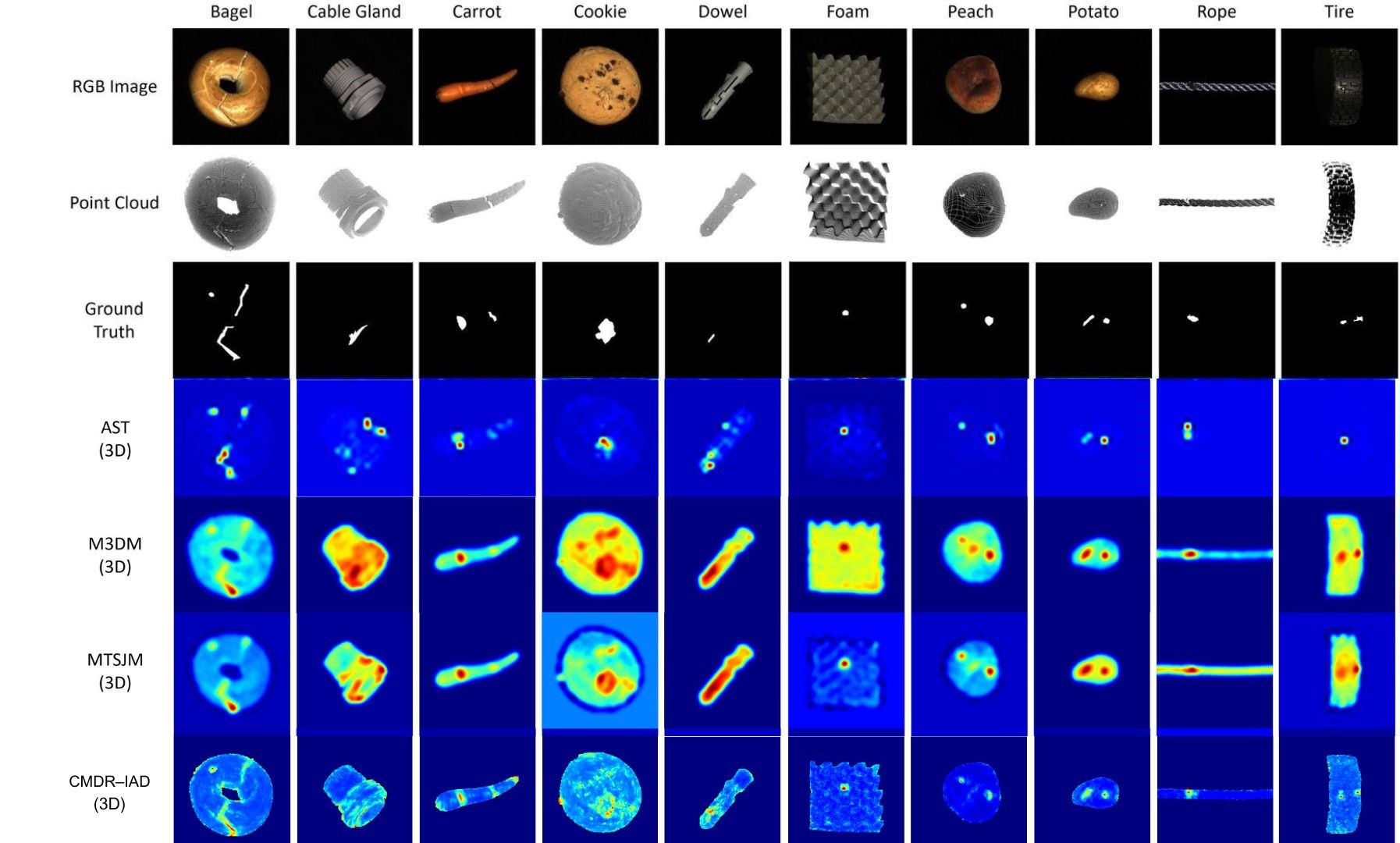}
    \caption{
Qualitative anomaly localization results on the MVTec 3D-AD dataset using \textbf{3D-only} point cloud features. From top to bottom, the rows show the RGB image, point cloud, ground truth, AST (3D), M3DM (3D), MTSJM (3D), and the proposed \textbf{CMDR--IAD} (3D). Warmer colors indicate higher anomaly scores, while cooler colors denote normal regions.
}
    \label{fig:Anomaly_map_mvtec_figure3d}
\end{figure}
\begin{figure}[H]
    \centering
    \includegraphics[width=1\textwidth]{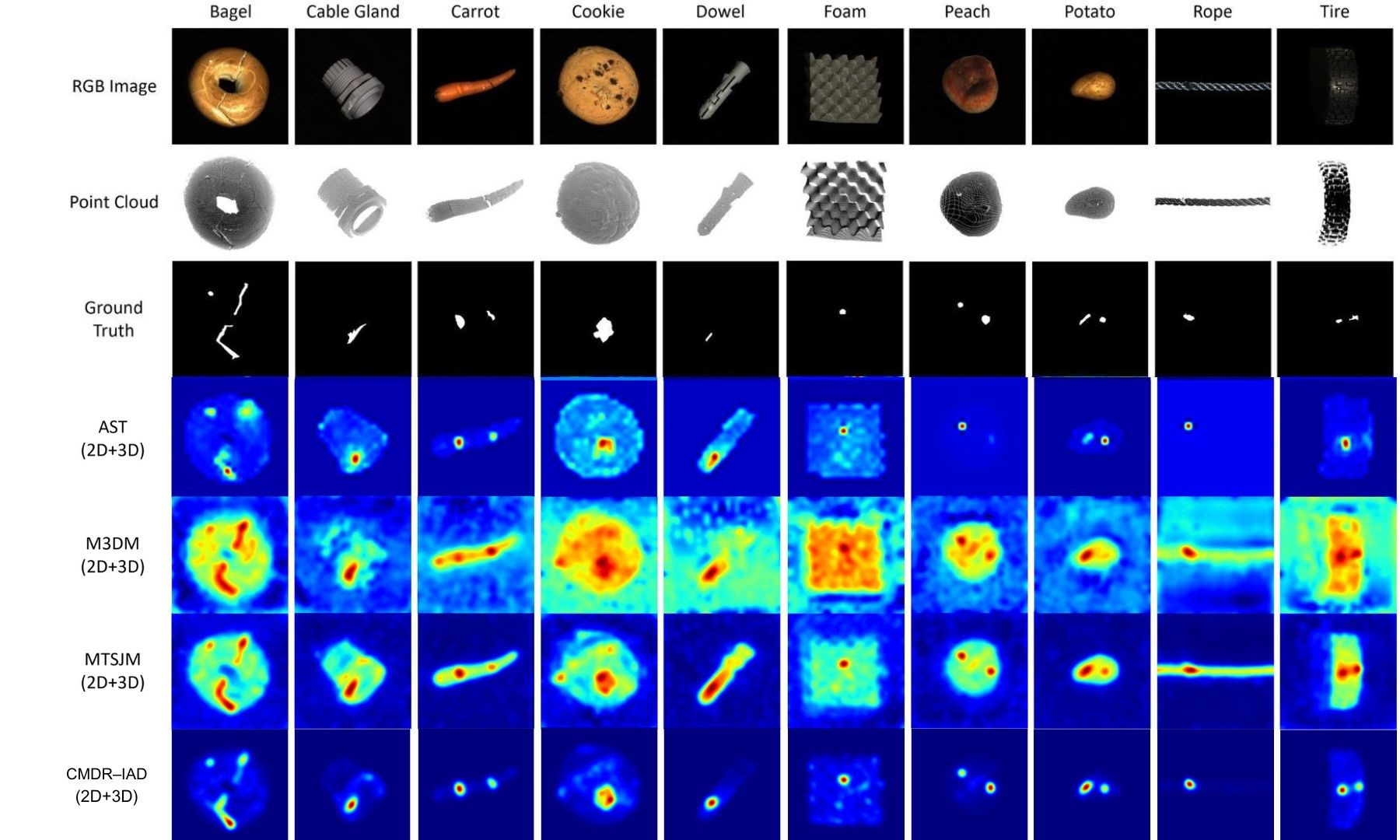}
    \caption{
Qualitative anomaly localization results on the MVTec 3D-AD dataset using \textbf{multimodal (2D+3D)} features. From top to bottom, the rows show the RGB image, point cloud, ground truth, AST (2D+3D), M3DM (2D+3D), MTSJM (2D+3D), and the proposed \textbf{CMDR--IAD} (2D+3D). Warmer colors indicate higher anomaly scores, while cooler colors denote normal regions.
}
    \label{fig:Anomaly_map_mvtec_figure2d+3d}
\end{figure}

\subsection{Results on Polyurethane Cutting Dataset (3D-Only)}
\subsubsection{Quantitative Results.}

\begin{table}[H]
\renewcommand{\arraystretch}{1.8} 
\centering
\caption{3D-only performance of \textbf{CMDR--IAD} on the Polyurethane Cutting dataset.
We report image-level AUROC (I-AUROC), point-level AUROC (P-AUROC), inference
speed measured in frames per second (FPS), and GPU memory footprint.}

\resizebox{0.7\linewidth}{!}{%
\begin{tabular}{lcccc}

\hline
\textbf { Metric } &  \textbf{Frame Rate (FPS)} &\textbf{ Memory (MB)} &\textbf{ I-AUROC } & \textbf{ P-AUROC } \\
\hline
\textbf{{CMDR--IAD} (Ours)} & 24.63 &465.68& 92.6 & 92.5  \\
\hline
\end{tabular}%
}
\label{tab:polyurethane_results}
\end{table}

\noindent Table~\ref{tab:polyurethane_results} reports the performance of \textbf{CMDR--IAD} on the
Polyurethane Cutting dataset using the 3D-only reconstruction branch. With the
Isolation Forest detector (low contamination) and a chunk size of 9216 points,
\textbf{CMDR--IAD} achieves an I-AUROC of 92.6 and a P-AUROC of 92.5, demonstrating strong
image-level anomaly detection and precise localization of geometric defects.
These results confirm that the proposed 3D reconstruction pathway effectively
models surface geometry and reliably detects subtle cutting irregularities even
without multimodal input.
\subsubsection{Qualitative Visualization.}

Fig.~\ref{fig:polyurethane_vis} illustrates qualitative anomaly localization results on representative samples from the polyurethane cuts dataset. Only 3D point cloud data is used during inference. Warmer colors correspond to higher anomaly scores, while cooler colors indicate normal regions.

\noindent As shown in Fig.~\ref{fig:polyurethane_vis}, \textbf{CMDR--IAD} accurately highlights surface cuts and material defects with minimal false responses on normal regions, despite the sparsity and noise of real industrial point clouds. These qualitative observations are consistent with the quantitative performance reported in Table~\ref{tab:polyurethane_results}, demonstrating the robustness of the proposed method under practical 3D-only inspection scenarios.

\begin{figure}[H]
    \centering
    \includegraphics[width=0.9\textwidth]{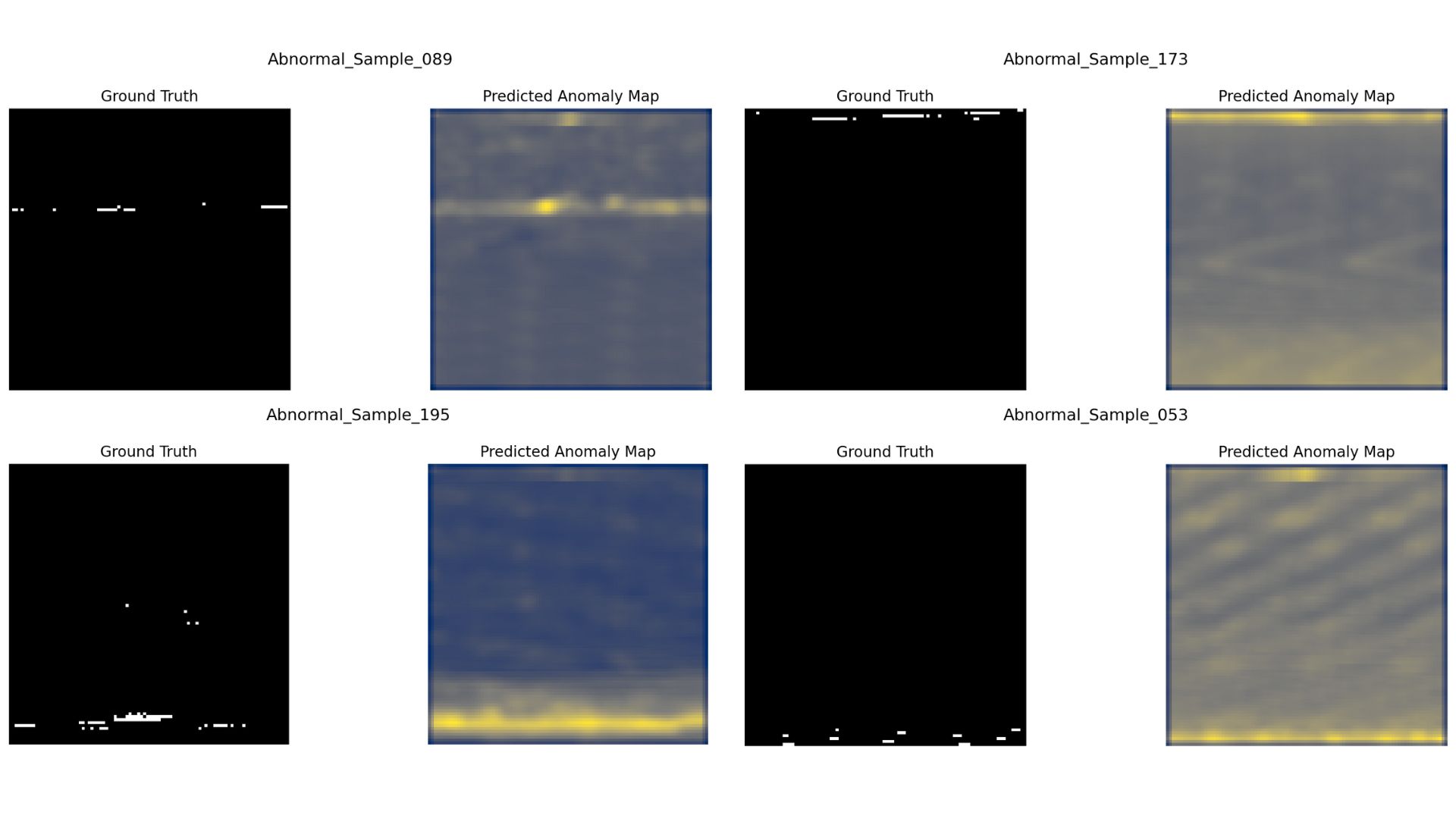}
    \caption{
Qualitative anomaly localization results on the polyurethane cuts dataset using \textbf{3D-only} point cloud data. For each sample, the ground truth mask and the corresponding predicted anomaly map are shown. Warmer colors indicate higher anomaly scores, while cooler colors denote normal regions.
}
    \label{fig:polyurethane_vis}
\end{figure}

\subsection{Ablation Study}

\subsubsection{Effect of dual-branch reconstruction and cross-modal mapping}

Table~\ref{tab:ablation_reconstruction_mapping} shows the impact of dual-branch reconstruction and cross-modal mapping on MVTec~3D-AD. Using either component alone already yields strong performance, with cross-modal mapping providing better localization under strict metrics. The full \textbf{CMDR--IAD} model consistently achieves the best results across all metrics, including AUPRO@30\% and AUPRO@1\%. This demonstrates that combining modality-specific reconstruction with cross-modal consistency provides complementary cues that are essential for robust anomaly detection and precise localization.

\begin{table}[ht]
\centering
\caption{Ablation study on the MVTec 3D-AD dataset evaluating the effect of dual-branch reconstruction and cross-modal mapping. Best results are in \textbf{bold}.}
\resizebox{0.7\linewidth}{!}{%
\begin{tabular}{lcccc}
\hline
\textbf{Configuration} & \textbf{I-AUROC} & \textbf{P-AUROC} & \textbf{AUPRO@30\%} & \textbf{AUPRO@1\%} \\
\hline
Dual Reconstruction only & 95.0 & 98.6 & 95.2 & 41.0  \\
Cross-Modal Mapping only & 95.4 &99.4 & 97.4 & 46.0 \\
\midrule
\textbf{{CMDR--IAD} (Ours)} & \textbf{97.3} & \textbf{99.6} & \textbf{97.6} & \textbf{46.5}  \\
\hline
\end{tabular}%
}
\label{tab:ablation_reconstruction_mapping}
\end{table}

\subsubsection{Effect of Multimodal Fusion Strategies}
\label{subsubsec:fusion_ablation}

We study the impact of different multimodal fusion strategies by evaluating
simplified anomaly scoring functions that combine cross-modal mapping
discrepancies and reconstruction errors in various ways.

\paragraph{\textbf{Ablation Variants}}
We define a set of alternative anomaly scoring functions $\Psi$ to isolate the
role of reliability gating, confidence weighting, and fusion design.

\textbf{Case 1. Gated Mapping Fusion}
\begin{equation}
\Psi_{C1}(x,y) = 
\alpha(x,y)\, d_{joint}(x,y)\, d_{rec}(x,y).
\label{eq:A1}
\end{equation}

\textbf{Case 2. Pure Multiplicative Fusion}
\begin{equation}
\Psi_{C2}(x,y) =
d^{2D}_{map}(x,y)\, d^{3D}_{map}(x,y)\,
d^{2D}_{rec}(x,y)\, d^{3D}_{rec}(x,y).
\label{eq:A2}
\end{equation}

\textbf{Case 3. Soft Adaptive Fusion (No Gating)}
\begin{align*}
d_{map}^{soft}(x,y) &=
\mathrm{softmax}\!\left(-d^{2D}_{map}(x,y), -d^{3D}_{map}(x,y)\right)
\cdot 
\begin{bmatrix}
d^{2D}_{map}(x,y) \\[2pt]
d^{3D}_{map}(x,y)
\end{bmatrix}, \\
d_{rec}^{soft}(x,y) &=
\mathrm{softmax}\!\left(-d^{2D}_{rec}(x,y), -d^{3D}_{rec}(x,y)\right)
\cdot
\begin{bmatrix}
d^{2D}_{rec}(x,y) \\[2pt]
d^{3D}_{rec}(x,y)
\end{bmatrix}.
\end{align*}
\begin{equation}
\Psi_{C3}(x,y) =
d_{map}^{soft}(x,y)\, d_{rec}^{soft}(x,y).
\label{eq:A3}
\end{equation}

\textbf{Case 4. Softmax Mapping + Gated Reconstruction}
\begin{align*}
d_{map}^{soft}(x,y) &=
\mathrm{softmax}\!\left(-d^{2D}_{map}(x,y), -d^{3D}_{map}(x,y)\right)
\cdot
\begin{bmatrix}
d^{2D}_{map}(x,y) \\[2pt]
d^{3D}_{map}(x,y)
\end{bmatrix}.
\end{align*}
\begin{equation}
\Psi_{C4}(x,y) =
d_{map}^{soft}(x,y)\,
\big[\alpha(x,y)\, d^{2D}_{rec}(x,y)\, d^{3D}_{rec}(x,y)\big].
\label{eq:A4}
\end{equation}

\textbf{Case 5. Dual Gated Joint Fusion}
\begin{align*}
d_{joint}(x,y) &= d^{2D}_{map}(x,y)\, d^{3D}_{map}(x,y).
\end{align*}
\begin{equation}
\Psi_{C5}(x,y) =
\big[\alpha(x,y)\, d_{joint}(x,y)\big]\,
\big[\alpha(x,y)\, d^{2D}_{rec}(x,y)\, d^{3D}_{rec}(x,y)\big].
\label{eq:A5}
\end{equation}

\textbf{Case 6. Uniform Averaging Baseline}
\begin{align*}
d_{avg}(x,y) &= 
\frac{
d^{2D}_{map}(x,y) + d^{3D}_{map}(x,y)
+ d^{2D}_{rec}(x,y) + d^{3D}_{rec}(x,y)
}{4}.
\end{align*}
\begin{equation}
\Psi_{C6}(x,y) = d_{avg}(x,y).
\label{eq:A6}
\end{equation}

\textbf{main paper.} \textbf{{CMDR--IAD} (Ours)}
\begin{align*}
w_2(x,y) &= \exp\!\big(-B\,d^{2D}_{rec}(x,y)\big), \\
w_3(x,y) &= \exp\!\big(-B\,d^{3D}_{rec}(x,y)\big). 
\end{align*}
\begin{equation}
\Psi_{full}(x,y)
=
\alpha(x,y)\, d_{joint}(x,y)\,
\frac{
w_2(x,y)\, d^{2D}_{rec}(x,y)
+
w_3(x,y)\, d^{3D}_{rec}(x,y)
}{
w_2(x,y) + w_3(x,y) + \varepsilon
}.
\label{eq:A0}
\end{equation}

\noindent Table~\ref{tab:ablation} shows that both cross-modal mapping and dual-branch
reconstruction contribute to anomaly detection, but naive fusion strategies are
suboptimal. Uniform averaging and ungated multiplicative fusion (Cases~2 and~6)
are sensitive to unreliable modalities and reduce localization accuracy.
Soft fusion without reliability gating (Case~3) further degrades performance.
In contrast, reliability-aware variants (Cases~1, 4, and~5) consistently
improve results. The full \textbf{CMDR--IAD} formulation achieves the best overall
performance by jointly leveraging reliability-gated cross-modal discrepancies
and confidence-weighted reconstruction errors, validating the effectiveness of
the proposed fusion strategy.
Detailed per-class ablation results for all fusion variants are provided in Appendix~A (Table~\ref{tab:appendix_ablation_per_class}).

\begin{table}[t]
\centering
\caption{Ablation study of multimodal anomaly scoring variants$\Psi_{C1}$--$\Psi_{C6}$ (Eqs.~\ref{eq:A1}--\ref{eq:A6}) compared to the full \textbf{CMDR--IAD} model (Eq.~\ref{eq:A0}) on the MVTec 3D--AD dataset.}

\label{tab:ablation}

\resizebox{\linewidth}{!}{
\begin{tabular}{lcccc}
\toprule
\textbf{Fusion Variant} & \textbf{I-AUROC} & \textbf{P-AUROC} & \textbf{AUPRO@30\%} & \textbf{AUPRO@1\%} \\
\midrule

Case 1: Gated Mapping Fusion (Eq.~\ref{eq:A1})
& 96.7 & 99.5 & 97.4 & 46.3 \\

Case 2: Pure Multiplicative (Eq.~\ref{eq:A2})
& 97.1 & 99.5 & 97.6 & 46.4 \\

Case 3: Soft Adaptive (Eq.~\ref{eq:A3})
& 96.7 & 99.2 & 96.8 & 45.0 \\

Case 4: Softmax + Gated Rec (Eq.~\ref{eq:A4})
& 96.6 & 99.5 & 97.4 & 46.1 \\

Case 5: Dual Gated Fusion (Eq.~\ref{eq:A5})
& {97.0} & {99.6} & {97.7} & {46.5} \\

Case 6: Uniform Avg. (Eq.~\ref{eq:A6})
& 95.0 & 98.7 & 95.7 & 42.5 \\

\midrule
\textbf{{CMDR--IAD} (Ours) (Eq.~\ref{eq:A0})}
& \textbf{97.3} & \textbf{99.6} & \textbf{97.6} & \textbf{46.5} \\

\bottomrule
\end{tabular}
}
\end{table}

\subsubsection{Effect of Outlier Detectors in the Polyurethane Preprocessing Pipeline}

Prior to chunking, each point cloud is filtered using an unsupervised outlier
detector. We evaluate Local Outlier Factor (LOF) and Isolation Forest (ISO)
under two contamination levels: 0.0001 (low) and 0.00015 (high). These values
reflect the extremely sparse nature of cutting defects in polyurethane scans,
with the higher setting being slightly more permissive.

\noindent LOF identifies points with locally inconsistent density, making it sensitive to
isolated irregularities but less stable on surfaces with varying point density.
In contrast, ISO isolates points via random partitioning and is more robust to
global geometric variability. After filtering, scans are divided into sequential
chunks of 4096 or 9216 points, where larger chunks preserve more geometric
context and improve decision consistency.

\noindent Table~\ref{tab:polyurethane_ablation} shows that LOF achieves high image-level
AUROC on large chunks but suffers from unstable localization. The hybrid
strategy (LOF $\cup$ ISO) increases coverage but introduces noise, reducing
robustness. ISO with low contamination and a chunk size of 9216 provides the
best balance between global detection and point-level localization
(92.6 I-AUROC, 92.5 P-AUROC), and is therefore adopted as the final
\textbf{CMDR--IAD} preprocessing configuration.

\begin{table}[t]
\centering
\caption{Ablation study of outlier detectors for polyurethane preprocessing,
comparing LOF, Isolation Forest (ISO), and a hybrid strategy across
contamination levels and chunk sizes. The configuration used in the final
\textbf{CMDR--AD} pipeline is highlighted.}
\label{tab:polyurethane_ablation}

\resizebox{0.60\linewidth}{!}{
\begin{tabular}{c|cc|c|c}
\toprule
Detector & Sample Size & Level & \textbf{I-AUROC} & \textbf{P-AUROC} \\
\midrule
\multirow{4}{*}{LOF} 
    & \multirow{2}{*}{4096} & low  & 80.1 & 94.6 \\
    &                       & high & 80.2 & 85.6 \\
\cline{2-5}
    & \multirow{2}{*}{9216} & low  & 99.5 & 80.3 \\
    &                       & high & 94.4 & 79.6 \\
\midrule
\multirow{4}{*}{ISO}
    & \multirow{2}{*}{4096} & low  & 87.5 & 87.4 \\
    &                       & high & 86.3 & 84.6 \\
\cline{2-5}
    & \multirow{2}{*}{9216} & low  
        & \textbf{92.6} & \textbf{92.5} \textbf{{(CMDR--AD)} Ours} \\
    &                       & high & 85.8 & 90.3 \\
\midrule
Hybrid (LOF $\cup$ ISO) & 9216 & low & 89.9 & 91.8 \\
\bottomrule
\end{tabular}
}
\end{table}

\section{Conclusion and Limitations}

In this work, we introduced \textbf{CMDR--IAD}, a multimodal anomaly detection framework that combines cross-modal feature mapping with dual-branch 2D--3D reconstruction. By independently modeling appearance and geometric patterns while enforcing bidirectional feature consistency, \textbf{CMDR--IAD} achieves robust anomaly localization across challenging industrial settings. The proposed adaptive fusion strategy integrates reconstruction deviations and cross-modal discrepancies through reliability weighting and contrast-aware gating, enabling stable predictions even in the presence of sparse, noisy, or incomplete 3D measurements. Experiments on the MVTec 3D--AD benchmark demonstrate that \textbf{CMDR--IAD}
achieves state-of-the-art performance with competitive inference speed and memory usage. Evaluation on a real polyurethane cutting dataset further confirms the generalization ability of the 3D-only variant and its suitability for geometry-dominant inspection tasks.

\noindent Despite its strengths, \textbf{CMDR--IAD} relies on pretrained feature extractors
and requires aligned RGB--3D data when cross-modal mapping is employed.
Furthermore, detecting anomalies with extremely subtle appearances or geometric
variations remains challenging. Future work will investigate addressing these
limitations through improved pretraining strategies and more effective fusion
mechanisms for capturing fine-grained cross-modal interactions.


\section*{CRediT authorship contribution statement}
Radia Daci: Conceptualization, Methodology, Software, Writing – Original Draft.  
Vito Reno: Funding Acquisition, Project Administration.
Cosimo Patruno: Data Curation, Resources.  
Angelo Cardellicchio: Resources, Data Curation.  
Abdelmalik Taleb-Ahmed:  Validation, Formal Analysis, Investigation.  
Marco Leo: Validation, Writing – Review \& Editing.
Cosimo Distante: Supervision, Methodology, Writing – Review \& Editing.

\section*{Declaration of competing interest}

The authors declare that they have no known competing financial interests or personal relationships that could have appeared to
influence the work reported in this paper.

\section*{Acknowledgment}
MO.RO.S.A.I. Mobile Robotic System with Artificial Intelligence Prog.n. F/310001/05/X56;
Programma di investimenti in attività di ricerca e sviluppo inerente alla linea di intervento del Programma
“Orizzonte Europa” “Intelligenza artificiale e robotica”; BANDO MISE - ACCORDI PER L'INNOVAZIONE (DD
18/03/2022 - DM 31/12/2021) approvato con Decreto di Concessione D.D. MIMIT n.0001207 13-04-2023;
CUP: B89J23001000005.

The authors thank Mr. Arturo Argentieri from CNR-ISASI, Italy, for his technical contribution to the multi-GPU computing facilities. }

\appendix
\section{Additional experiment results on MVTec 3D-AD}

Although P-AUROC is commonly reported, it provides limited discrimination for evaluating multimodal information fusion on MVTec 3D-AD, as most methods achieve consistently high and closely clustered P-AUROC values; the corresponding results are therefore reported in Appendix~A  Table~\ref{tab:pauroc_mvtec3dad} for completeness.

\begin{table*}[ht]
\centering
\caption{P-AUROC (\%) for anomaly detection on all classes of the MVTec 3D-AD dataset.
Best results are shown in bold and second-best results are underlined.
Our method achieves the best performance in the 3D and 2D+3D settings, and the second-best performance in the 2D setting.}

\label{tab:pauroc_mvtec3dad}

\resizebox{\linewidth}{!}{
\begin{tabular}{lccccccccccc}
\toprule
Method & Bagel & CableGland & Carrot & Cookie & Dowel & Foam & Peach & Potato & Rope & Tire & Mean \\
\midrule
\multicolumn{12}{c}{\textbf{2D}} \\
\midrule
PatchCore (2022) \cite{roth2022towards}          & 98.3 & 98.4 & 98.0 & 97.4 & 97.2 & 84.9 & 97.6 & 98.3 & 98.7 & 97.7 & 96.7 \\
M3DM (2023) \cite{wang2023multimodal}               & 99.2 & 99.0 & 99.4 & 97.7 & 98.3 & 95.5 & 99.4 & 99.0 & 99.5 & 99.4 & \textbf{98.7} \\
CMDIAD (2025) \cite{sui2025incomplete}      & 99.2 & 99.3 & 99.4 & 97.7 & 98.3 & 95.6 & 99.3 & 99.0 & 99.5 & 99.4 & \textbf{98.7} \\
\textbf{{CMDR--IAD} (Ours)}&  
98.6 & 98.8 & 94.8 & 99.3 & 99.4 & 94.0 & 97.2 & 99.6 & 99.6 & 99.2 &\underline{ 98.1} \\

\midrule

\multicolumn{12}{c}{\textbf{3D}} \\
\midrule
FPFH (2023) \cite{horwitz2023back}               & 99.4 & 96.6 & 99.9 & 94.6 & 96.6 & 92.7 & 99.6 & 99.9 & 99.6 & 99.0 & \underline{97.8} \\
M3DM (2023)  \cite{wang2023multimodal}               & 98.1 & 94.9 & 99.7 & 93.2 & 95.9 & 92.5 & 98.9 & 99.5 & 99.4 & 98.1 & 97.0 \\
CMDIAD (2025) \cite{sui2025incomplete}       & 98.3 & 95.0 & 99.7 & 93.2 & 95.9 & 94.0 & 99.1 & 99.6 & 99.4 & 98.4 & 97.3 \\
\textbf{{CMDR--IAD} (Ours)}
& 95.3 & 99.8 & 93.5 & 99.2 & 99.7 & 96.3 & 97.2 & 99.6 & 99.8 & 99.2 & \textbf{98.0} \\

\midrule
\multicolumn{12}{c}{\textbf{2D+3D}} \\
\midrule

PatchCore (2022) \cite{roth2022towards}          & \underline{99.6} & \underline{99.2} & \underline{99.7} & \textbf{99.4} & 98.1 & 97.4 & \underline{99.6} &  \underline{99.8} & 99.4 & 99.5 & \underline{99.2} \\
M3DM (2023) \cite{wang2023multimodal}& 99.5 & \textbf{99.3 }& \underline{99.7} & \underline{98.5} & \underline{98.5} & 98.4 & \underline{99.6} & 99.4 & \underline{99.7} & \underline{99.6 }& \underline{99.2} \\
CMDIAD (2025) \cite{sui2025incomplete}      & 99.5 & \textbf{99.3} & 99.6 & 97.6 & 98.4 & \underline{98.8} & \underline{99.6} & {99.5 }& \underline{99.7 }& \underline{99.6} & \underline{99.2} \\

\textbf{{CMDR--IAD} (Ours)} &
\textbf{99.7} & \textbf{99.3} & \textbf{99.9} & {98.3} & \textbf{99.2} & \textbf{99.6} & \textbf{99.9} & \textbf{99.9} &  \textbf{99.9} &  \textbf{99.8} & \textbf{99.6} \\

\bottomrule
\end{tabular}
}

\end{table*}
\section{Additional Ablation Results on the MVTec 3D-AD Dataset}
This section reports detailed per-class ablation results corresponding to the multimodal fusion variants discussed in Section~\ref{subsubsec:fusion_ablation}.
The results complement Table~\ref{tab:ablation} by providing class-wise performance on the MVTec 3D-AD dataset.

\noindent Consistent with the main results, the full \textbf{CMDR--IAD} model achieves the strongest
overall performance across most object classes and evaluation metrics.
Performance degradations observed in the ablation variants further highlight
the importance of reliability-aware multimodal fusion.

\begin{table*}[t]
\centering
\caption{\textbf{Per-class ablation study for all multimodal fusion variants.}
Columns correspond to the 10 MVTec 3D-AD classes, while rows report four evaluation metrics
for each ablation case defined in Eqs.~(\ref{eq:A1})–(\ref{eq:A6}) and the full model (Eq.~\ref{eq:A0}).
The rightmost column reports the mean performance across all classes.}

\label{tab:appendix_ablation_per_class}

\resizebox{\textwidth}{!}{
\begin{tabular}{l|c|cccccccccc|c}
\toprule
\textbf{Case} & \textbf{Metric} &
\textbf{Bagel} & \textbf{CableGland} & \textbf{Carrot} & \textbf{Cookie} &
\textbf{Dowel} & \textbf{Foam} & \textbf{Peach} & \textbf{Potato} &
\textbf{Rope} & \textbf{Tire} & \textbf{Mean} \\
\midrule

\multirow{4}{*}{\textbf{Case 1} (Eq.~\ref{eq:A1})}
& I-AUROC  & 99.5 & 91.7 & 98.6 & 99.7 & 98.8 & 92.1 & 99.4 & 91.7 & 98.8 & 96.5 & \textbf{96.7} \\
& P-AUROC  & 99.7 & 99.3 & 99.8 & 98.0 & 99.0 & 99.5 & 99.9 & 99.8 & 99.8 & 99.8 & \textbf{99.5} \\
& AUPRO@30 & 97.9 & 97.3 & 98.2 & 95.4 & 96.0 & 97.0 & 98.2 & 98.3 & 97.9 & 98.2 & \textbf{97.4} \\
& AUPRO@1  & 46.9 & 44.2 & 48.5 & 46.5 & 40.8 & 43.2 & 48.6 & 48.8 & 47.2 & 48.4 & \textbf{46.3} \\
\midrule

\multirow{4}{*}{\textbf{Case 2} (Eq.~\ref{eq:A2})}
& I-AUROC  & 99.7 & 87.2 & 98.6 & 99.7 & 98.8 & 93.8 & 99.6 & 96.0 & 99.0 & 98.7 & \textbf{97.1} \\
& P-AUROC  & 99.7 & 98.8 & 99.8 & 98.6 & 99.2 & 99.7 & 99.9 & 99.9 & 99.9 & 99.8 & \textbf{99.5} \\
& AUPRO@30 & 98.0 & 96.1 & 98.2 & 96.3 & 96.6 & 97.7 & 98.3 & 98.3 & 98.1 & 98.3 & \textbf{97.6} \\
& AUPRO@1  & 47.1 & 39.7 & 48.5 & 46.2 & 42.0 & 44.9 & 49.2 & 49.6 & 47.5 & 49.2 & \textbf{46.4} \\
\midrule

\multirow{4}{*}{\textbf{Case 3} (Eq.~\ref{eq:A3})}
& I-AUROC  & 99.5 & 92.0 & 99.1 & 99.4 & 98.9 & 94.5 & 99.4 & 89.3 & 98.5 & 96.3 & \textbf{96.7} \\
& P-AUROC  & 99.6 & 99.3 & 99.8 & 96.7 & 98.2 & 99.5 & 99.9 & 99.8 & 99.8 & 99.8 & \textbf{99.2} \\
& AUPRO@30 & 97.6 & 96.9 & 98.2 & 93.3 & 93.4 & 97.1 & 98.2 & 98.1 & 97.3 & 98.2 & \textbf{96.8} \\
& AUPRO@1  & 45.5 & 43.8 & 48.1 & 44.8 & 35.7 & 43.3 & 48.9 & 46.8 & 45.5 & 47.9 & \textbf{45.0} \\
\midrule

\multirow{4}{*}{\textbf{Case 4} (Eq.~\ref{eq:A4})}
& I-AUROC  & 99.4 & 83.3 & 99.0 & 99.1 & 98.7 & 92.5 & 99.4 & 96.7 & 99.3 & 98.9 & \textbf{96.6} \\
& P-AUROC  & 99.7 & 98.7 & 99.8 & 98.3 & 98.9 & 99.7 & 99.9 & 99.9 & 99.9 & 99.8 & \textbf{99.5} \\
& AUPRO@30 & 97.9 & 95.5 & 98.2 & 96.0 & 95.7 & 97.8 & 98.3 & 98.3 & 98.0 & 98.3 & \textbf{97.4} \\
& AUPRO@1  & 47.1 & 38.4 & 48.5 & 46.8 & 39.8 & 46.2 & 49.2 & 49.1 & 47.2 & 49.2 & \textbf{46.1} \\
\midrule

\multirow{4}{*}{\textbf{Case 5} (Eq.~\ref{eq:A5})}
& I-AUROC  & 99.7 & 86.5 & 98.5 & 99.6 & 99.0 & 93.1 & 99.6 & 96.6 & 99.0 & 99.0 & \textbf{97.0} \\
& P-AUROC  & 99.7 & 98.9 & 99.8 & 99.0 & 99.4 & 99.7 & 99.9 & 99.9 & 99.9 & 99.8 & \textbf{99.6} \\
& AUPRO@30 & 97.9 & 96.2 & 98.2 & 96.9 & 97.1 & 97.5 & 98.3 & 98.3 & 98.1 & 98.3 & \textbf{97.7} \\
& AUPRO@1  & 47.0 & 39.9 & 48.6 & 46.4 & 43.0 & 44.3 & 49.7 & 49.4 & 47.7 & 49.0 & \textbf{46.5} \\
\midrule

\multirow{4}{*}{\textbf{Case 6} (Eq.~\ref{eq:A6})}
& I-AUROC  & 99.1 & 90.4 & 99.4 & 99.0 & 98.7 & 91.6 & 98.8 & 82.7 & 98.7 & 92.1 & \textbf{95.0} \\
& P-AUROC  & 99.1 & 99.1 & 99.7 & 94.8 & 97.2 & 99.2 & 99.8 & 99.5 & 99.6 & 99.5 & \textbf{98.7} \\
& AUPRO@30 & 96.6 & 95.9 & 97.9 & 90.2 & 90.5 & 96.3 & 98.2 & 97.6 & 96.0 & 97.5 & \textbf{95.7} \\
& AUPRO@1  & 43.5 & 42.2 & 46.3 & 42.9 & 30.2 & 41.1 & 48.7 & 43.2 & 41.5 & 45.7 & \textbf{42.5} \\
\midrule

\multirow{4}{*}{\textbf{CMDR--IAD (Ours)} (Eq.~\ref{eq:A0})}
& I-AUROC  & 99.6 & 93.0 & 98.6 & 99.8 & 99.1 & 93.6 & 99.6 & 93.1 & 98.7 & 97.5 & \textbf{97.3} \\
& P-AUROC  & 99.7 & 99.3 & 99.9 & 98.3 & 99.2 & 99.6 & 99.9 & 99.9 & 99.9 & 99.8 & \textbf{99.5} \\
& AUPRO@30 & 98.0 & 97.3 & 98.2 & 95.9 & 96.6 & 97.1 & 98.2 & 98.2 & 98.0 & 98.2 & \textbf{97.6} \\
& AUPRO@1  & 47.2 & 43.9 & 48.6 & 46.4 & 42.2 & 43.5 & 48.6 & 48.5 & 47.4 & 48.5 & \textbf{46.5} \\

\bottomrule
\end{tabular}
}
\end{table*}

\section{Additional Qualitative Results}

\begin{figure}[H]
    \centering
    \includegraphics[width=\textwidth]{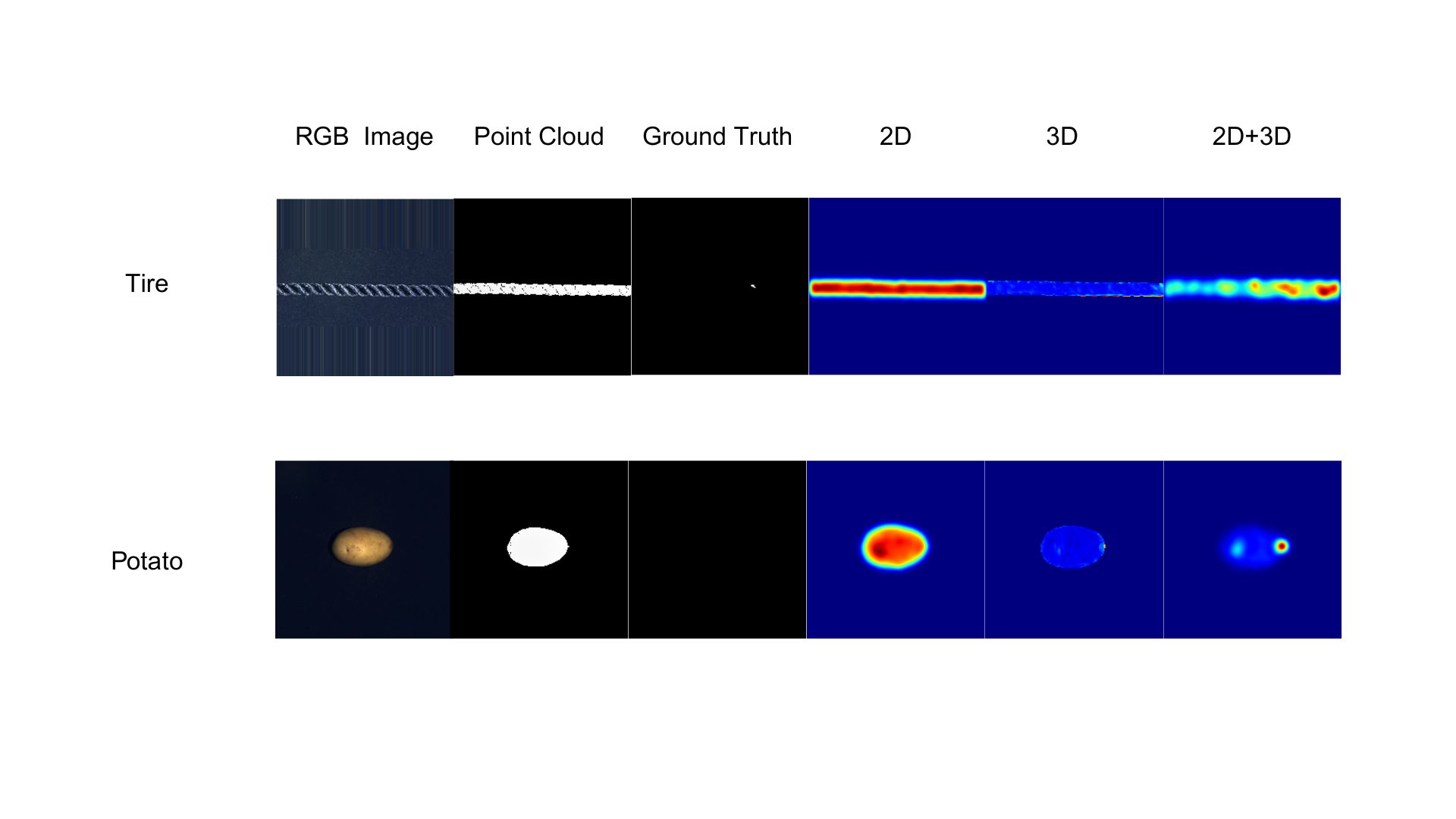}
    \caption{Failure case examples on the MVTec 3D-AD dataset.
    From left to right: RGB image, point cloud, ground truth, and anomaly maps produced by
    2D-only, 3D-only, and multimodal (2D+3D) configurations.}
    \label{fig:Anomaly_map_mvtec_failure}
\end{figure}

\begin{figure}[H]
     \centering
    \includegraphics[width=1\textwidth]{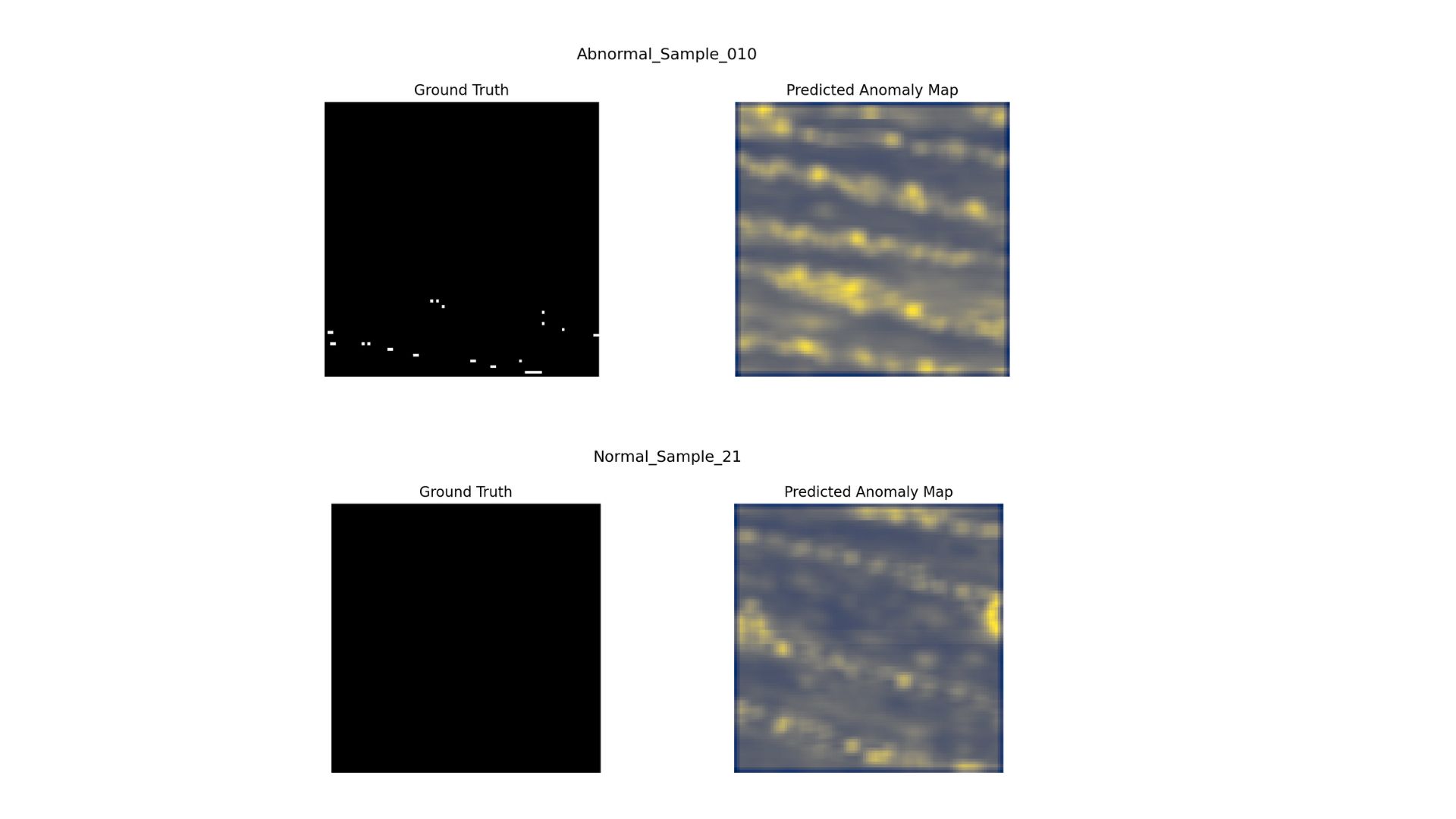}
    \caption{Examples of false positive and false negative predictions on the Polyurethane Cutting dataset.
Normal samples may exhibit elevated anomaly responses due to unseen appearance or geometric patterns,
while subtle defect regions may be weakly detected or missed in the predicted anomaly maps.}
    \label{fig:Anomaly_map_poly_failure}
\end{figure}

This section analyzes representative false positive and false negative samples that
affect the performance of the proposed \textbf{CMDR--IAD} method on the MVTec 3D-AD
and Polyurethane Cutting datasets.
These examples illustrate common failure modes caused by unseen normal patterns,
very small defect regions, and modality-specific limitations in 2D and 3D anomaly
detection, as shown in Figures~\ref{fig:Anomaly_map_mvtec_failure}
and~\ref{fig:Anomaly_map_poly_failure}.

\noindent As shown in Figure~\ref{fig:Anomaly_map_mvtec_failure}, false positives
mainly occur in normal regions that have not been observed during training,
resulting in elevated anomaly responses despite the absence of true defects.
False negatives are primarily associated with very small or low-contrast defect
regions, where anomaly signals may be dominated by background texture or geometric
noise.
In addition, as illustrated in Figure~\ref{fig:Anomaly_map_poly_failure},
modality-specific factors, such as point cloud acquisition noise in industrial
settings, can interfere with 3D-based detection of subtle geometric deformations.

\bibliographystyle{elsarticle-num}

\bibliography{references}

@String(AAAI  = {AAAI})

@article{bergmann2021mvtec,
  title={The mvtec 3d-ad dataset for unsupervised 3d anomaly detection and localization},
  author={Bergmann, Paul and Jin, Xin and Sattlegger, David and Steger, Carsten},
  journal={arXiv preprint arXiv:2112.09045},
  year={2021}
}

@inproceedings{costanzino2024multimodal,
  title={Multimodal industrial anomaly detection by crossmodal feature mapping},
  author={Costanzino, Alex and Ramirez, Pierluigi Zama and Lisanti, Giuseppe and Di Stefano, Luigi},
  booktitle={Proceedings of the IEEE/CVF Conference on Computer Vision and Pattern Recognition},
  pages={17234--17243},
  year={2024}
}

@article{sui2025incomplete,
  title={Incomplete multimodal industrial anomaly detection via cross-modal distillation},
  author={Sui, Wenbo and Lichau, Daniel and Lef{\`e}vre, Josselin and Phelippeau, Harold},
  journal={Information Fusion},
  pages={103572},
  year={2025},
  publisher={Elsevier}
}

@inproceedings{long2025revisiting,
  title={Revisiting multimodal fusion for 3D anomaly detection from an architectural perspective},
  author={Long, Kaifang and Xie, Guoyang and Ma, Lianbo and Liu, Jiaqi and Lu, Zhichao},
  booktitle={Proceedings of the AAAI Conference on Artificial Intelligence},
  volume={39},
  number={12},
  pages={12273--12281},
  year={2025}
}

@article{liu2025multimodal,
  title={A multimodal industrial anomaly detection method based on mask training and teacher--student joint memory},
  author={Liu, Yi and Zhang, Changsheng and Dong, Xingjun and Yang, Yufei},
  journal={Engineering Applications of Artificial Intelligence},
  volume={161},
  pages={112299},
  year={2025},
  publisher={Elsevier}
}

@article{bougaham2024composite,
  title={Composite score for anomaly detection in imbalanced real-world industrial dataset},
  author={Bougaham, Arnaud and El Adoui, Mohammed and Linden, Isabelle and Fr{\'e}nay, Beno{\^\i}t},
  journal={Machine Learning},
  volume={113},
  number={7},
  pages={4381--4406},
  year={2024},
  publisher={Springer}
}

@inproceedings{horwitz2023back,
  title={Back to the feature: classical 3d features are (almost) all you need for 3d anomaly detection},
  author={Horwitz, Eliahu and Hoshen, Yedid},
  booktitle={Proceedings of the IEEE/CVF Conference on Computer Vision and Pattern Recognition},
  pages={2968--2977},
  year={2023}
}

@inproceedings{rudolph2023asymmetric,
  title={Asymmetric student-teacher networks for industrial anomaly detection},
  author={Rudolph, Marco and Wehrbein, Tom and Rosenhahn, Bodo and Wandt, Bastian},
  booktitle={Proceedings of the IEEE/CVF winter conference on applications of computer vision},
  pages={2592--2602},
  year={2023}
}

@inproceedings{wang2023multimodal,
  title={Multimodal industrial anomaly detection via hybrid fusion},
  author={Wang, Yue and Peng, Jinlong and Zhang, Jiangning and Yi, Ran and Wang, Yabiao and Wang, Chengjie},
  booktitle={Proceedings of the IEEE/CVF conference on computer vision and pattern recognition},
  pages={8032--8041},
  year={2023}
}

@inproceedings{roth2022towards,
  title={Towards total recall in industrial anomaly detection},
  author={Roth, Karsten and Pemula, Latha and Zepeda, Joaquin and Sch{\"o}lkopf, Bernhard and Brox, Thomas and Gehler, Peter},
  booktitle={Proceedings of the IEEE/CVF conference on computer vision and pattern recognition},
  pages={14318--14328},
  year={2022}
}

@inproceedings{bergmann2023anomaly,
  title={Anomaly detection in 3d point clouds using deep geometric descriptors},
  author={Bergmann, Paul and Sattlegger, David},
  booktitle={Proceedings of the IEEE/CVF Winter Conference on Applications of Computer Vision},
  pages={2613--2623},
  year={2023}
}

@article{lin2024back,
  title={Back to the Metrics: Exploration of Distance Metrics in Anomaly Detection},
  author={Lin, Yujing and Li, Xiaoqiang},
  year={2024},
  publisher={Preprints}
}

@article{pang2023masked,
  title={Masked autoencoders for 3d point cloud self-supervised learning},
  author={Pang, Yatian and Tay, Eng Hock Francis and Yuan, Li and Chen, Zhenghua},
  journal={World Scientific Annual Review of Artificial Intelligence},
  volume={1},
  pages={2440001},
  year={2023},
  publisher={World Scientific}
}

@article{liu2024deep,
  title={Deep industrial image anomaly detection: A survey},
  author={Liu, Jiaqi and Xie, Guoyang and Wang, Jinbao and Li, Shangnian and Wang, Chengjie and Zheng, Feng and Jin, Yaochu},
  journal={Machine Intelligence Research},
  volume={21},
  number={1},
  pages={104--135},
  year={2024},
  publisher={Springer}
}

@article{angiulli2023latent,
  title={Latent O ut: an unsupervised deep anomaly detection approach exploiting latent space distribution},
  author={Angiulli, Fabrizio and Fassetti, Fabio and Ferragina, Luca},
  journal={Machine Learning},
  volume={112},
  number={11},
  pages={4323--4349},
  year={2023},
  publisher={Springer}
}

@article{wang2020image,
  title={Image anomaly detection using normal data only by latent space resampling},
  author={Wang, Lu and Zhang, Dongkai and Guo, Jiahao and Han, Yuexing},
  journal={Applied Sciences},
  volume={10},
  number={23},
  pages={8660},
  year={2020},
  publisher={MDPI}
}

@article{teng2022unsupervised,
  title={Unsupervised visual defect detection with score-based generative model},
  author={Teng, Yapeng and Li, Haoyang and Cai, Fuzhen and Shao, Ming and Xia, Siyu},
  journal={arXiv preprint arXiv:2211.16092},
  year={2022}
}

@inproceedings{yan2021learning,
  title={Learning semantic context from normal samples for unsupervised anomaly detection},
  author={Yan, Xudong and Zhang, Huaidong and Xu, Xuemiao and Hu, Xiaowei and Heng, Pheng-Ann},
  booktitle={Proceedings of the AAAI conference on artificial intelligence},
  volume={35},
  number={4},
  pages={3110--3118},
  year={2021}
}

@inproceedings{wyatt2022anoddpm,
  title={Anoddpm: Anomaly detection with denoising diffusion probabilistic models using simplex noise},
  author={Wyatt, Julian and Leach, Adam and Schmon, Sebastian M and Willcocks, Chris G},
  booktitle={Proceedings of the IEEE/CVF conference on computer vision and pattern recognition},
  pages={650--656},
  year={2022}
}

@article{bergmann2018improving,
  title={Improving unsupervised defect segmentation by applying structural similarity to autoencoders},
  author={Bergmann, Paul and L{\"o}we, Sindy and Fauser, Michael and Sattlegger, David and Steger, Carsten},
  journal={arXiv preprint arXiv:1807.02011},
  year={2018}
}

@inproceedings{hou2021divide,
  title={Divide-and-assemble: Learning block-wise memory for unsupervised anomaly detection},
  author={Hou, Jinlei and Zhang, Yingying and Zhong, Qiaoyong and Xie, Di and Pu, Shiliang and Zhou, Hong},
  booktitle={Proceedings of the IEEE/CVF International Conference on Computer Vision},
  pages={8791--8800},
  year={2021}
}

@inproceedings{ristea2022self,
  title={Self-supervised predictive convolutional attentive block for anomaly detection},
  author={Ristea, Nicolae-C{\u{a}}t{\u{a}}lin and Madan, Neelu and Ionescu, Radu Tudor and Nasrollahi, Kamal and Khan, Fahad Shahbaz and Moeslund, Thomas B and Shah, Mubarak},
  booktitle={Proceedings of the IEEE/CVF conference on computer vision and pattern recognition},
  pages={13576--13586},
  year={2022}
}

@inproceedings{zavrtanik2021draem,
  title={Draem-a discriminatively trained reconstruction embedding for surface anomaly detection},
  author={Zavrtanik, Vitjan and Kristan, Matej and Sko{\v{c}}aj, Danijel},
  booktitle={Proceedings of the IEEE/CVF international conference on computer vision},
  pages={8330--8339},
  year={2021}
}

@inproceedings{pirnay2022inpainting,
  title={Inpainting transformer for anomaly detection},
  author={Pirnay, Jonathan and Chai, Keng},
  booktitle={International Conference on Image Analysis and Processing},
  pages={394--406},
  year={2022},
  organization={Springer}
}

@inproceedings{li2021cutpaste,
  title={Cutpaste: Self-supervised learning for anomaly detection and localization},
  author={Li, Chun-Liang and Sohn, Kihyuk and Yoon, Jinsung and Pfister, Tomas},
  booktitle={Proceedings of the IEEE/CVF conference on computer vision and pattern recognition},
  pages={9664--9674},
  year={2021}
}

@inproceedings{chiu2023self,
  title={Self-supervised normalizing flows for image anomaly detection and localization},
  author={Chiu, Li-Ling and Lai, Shang-Hong},
  booktitle={Proceedings of the IEEE/CVF conference on computer vision and pattern recognition},
  pages={2927--2936},
  year={2023}
}

@inproceedings{gudovskiy2022cflow,
  title={Cflow-ad: Real-time unsupervised anomaly detection with localization via conditional normalizing flows},
  author={Gudovskiy, Denis and Ishizaka, Shun and Kozuka, Kazuki},
  booktitle={Proceedings of the IEEE/CVF winter conference on applications of computer vision},
  pages={98--107},
  year={2022}
}

@inproceedings{salehi2021multiresolution,
  title={Multiresolution knowledge distillation for anomaly detection},
  author={Salehi, Mohammadreza and Sadjadi, Niousha and Baselizadeh, Soroosh and Rohban, Mohammad H and Rabiee, Hamid R},
  booktitle={Proceedings of the IEEE/CVF conference on computer vision and pattern recognition},
  pages={14902--14912},
  year={2021}
}

@article{zhang2021anomaly,
  title={Anomaly detection using improved deep SVDD model with data structure preservation},
  author={Zhang, Zheng and Deng, Xiaogang},
  journal={Pattern Recognition Letters},
  volume={148},
  pages={1--6},
  year={2021},
  publisher={Elsevier}
}

@article{yang2020dfr,
  title={Dfr: Deep feature reconstruction for unsupervised anomaly segmentation},
  author={Yang, Jie and Shi, Yong and Qi, Zhiquan},
  journal={arXiv preprint arXiv:2012.07122},
  year={2020}
}

@inproceedings{caron2021emerging,
  title={Emerging properties in self-supervised vision transformers},
  author={Caron, Mathilde and Touvron, Hugo and Misra, Ishan and J{\'e}gou, Herv{\'e} and Mairal, Julien and Bojanowski, Piotr and Joulin, Armand},
  booktitle={Proceedings of the IEEE/CVF international conference on computer vision},
  pages={9650--9660},
  year={2021}
}

@inproceedings{he2022masked,
  title={Masked autoencoders are scalable vision learners},
  author={He, Kaiming and Chen, Xinlei and Xie, Saining and Li, Yanghao and Doll{\'a}r, Piotr and Girshick, Ross},
  booktitle={Proceedings of the IEEE/CVF conference on computer vision and pattern recognition},
  pages={16000--16009},
  year={2022}
}

@article{oquab2023dinov2,
  title={Dinov2: Learning robust visual features without supervision},
  author={Oquab, Maxime and Darcet, Timoth{\'e}e and Moutakanni, Th{\'e}o and Vo, Huy and Szafraniec, Marc and Khalidov, Vasil and Fernandez, Pierre and Haziza, Daniel and Massa, Francisco and El-Nouby, Alaaeldin and others},
  journal={arXiv preprint arXiv:2304.07193},
  year={2023}
}

@article{bergman2020deep,
  title={Deep nearest neighbor anomaly detection},
  author={Bergman, Liron and Cohen, Niv and Hoshen, Yedid},
  journal={arXiv preprint arXiv:2002.10445},
  year={2020}
}

@article{lee2022cfa,
  title={Cfa: Coupled-hypersphere-based feature adaptation for target-oriented anomaly localization},
  author={Lee, Sungwook and Lee, Seunghyun and Song, Byung Cheol},
  journal={IEEE Access},
  volume={10},
  pages={78446--78454},
  year={2022},
  publisher={IEEE}
}

@inproceedings{zou2022spot, 
  title={Spot-the-difference self-supervised pre-training for anomaly detection and segmentation},
  author={Zou, Yang and Jeong, Jongheon and Pemula, Latha and Zhang, Dongqing and Dabeer, Onkar},
  booktitle={European conference on computer vision},
  pages={392--408},
  year={2022},
  organization={Springer}
}

@article{qin2023teacher,
  title={Teacher--student network for 3D point cloud anomaly detection with few normal samples},
  author={Qin, Jianjian and Gu, Chunzhi and Yu, Jun and Zhang, Chao},
  journal={Expert Systems with Applications},
  volume={228},
  pages={120371},
  year={2023},
  publisher={Elsevier}
}

@inproceedings{liu2008isolation,
  title={Isolation forest},
  author={Liu, Fei Tony and Ting, Kai Ming and Zhou, Zhi-Hua},
  booktitle={2008 eighth ieee international conference on data mining},
  pages={413--422},
  year={2008},
  organization={IEEE}
}

@article{hackel2017semantic3d,
  title={Semantic3d. net: A new large-scale point cloud classification benchmark},
  author={Hackel, Timo and Savinov, Nikolay and Ladicky, Lubor and Wegner, Jan D and Schindler, Konrad and Pollefeys, Marc},
  journal={arXiv preprint arXiv:1704.03847},
  year={2017}
}

@incollection{lecun2012efficient,
  title     = {Efficient BackProp},
  author    = {LeCun, Yann A. and Bottou, L{\'e}on and Orr, Genevieve B. and M{\"u}ller, Klaus-Robert},
  booktitle = {Neural Networks: Tricks of the Trade},
  publisher = {Springer},
  year      = {2012}
}

@article{kingma2014adam,
  title={Adam: A method for stochastic optimization},
  author={Kingma, Diederik P},
  journal={arXiv preprint arXiv:1412.6980},
  year={2014}
}

@article{loshchilov2017decoupled,
  title={Decoupled weight decay regularization},
  author={Loshchilov, Ilya and Hutter, Frank},
  journal={arXiv preprint arXiv:1711.05101},
  year={2017}
}

@inproceedings{deng2009imagenet,
  title={Imagenet: A large-scale hierarchical image database},
  author={Deng, Jia and Dong, Wei and Socher, Richard and Li, Li-Jia and Li, Kai and Fei-Fei, Li},
  booktitle={2009 IEEE conference on computer vision and pattern recognition},
  pages={248--255},
  year={2009},
  organization={Ieee}
}

@article{chang2015shapenet,
  title={Shapenet: An information-rich 3d model repository},
  author={Chang, Angel X and Funkhouser, Thomas and Guibas, Leonidas and Hanrahan, Pat and Huang, Qixing and Li, Zimo and Savarese, Silvio and Savva, Manolis and Song, Shuran and Su, Hao and others},
  journal={arXiv preprint arXiv:1512.03012},
  year={2015}
}

@article{zhang2026dynamic,
  title={Dynamic background-guided asymmetric knowledge distillation network for 3D defect detection},
  author={Zhang, Fan and Hu, Wenlong and Wang, Yun and others},
  journal={Journal of Manufacturing Processes},
  volume={160},
  pages={185--199},
  year={2026},
  publisher={Elsevier}
}

\end{document}